\documentclass{article} 
\usepackage{iclr2023_conference,times}

\usepackage{amsmath,amsfonts,bm}

\def\eqref#1{equation~\ref{#1}}

\def\1{\bm{1}}

\DeclareMathAlphabet{\mathsfit}{\encodingdefault}{\sfdefault}{m}{sl}
\SetMathAlphabet{\mathsfit}{bold}{\encodingdefault}{\sfdefault}{bx}{n}

\newcommand{\E}{\mathbb{E}}

\newcommand{\ind}[1]{[[#1]]}
\usepackage{color}
\usepackage[hidelinks]{hyperref}
\usepackage{url}
\usepackage{natbib}

\usepackage{caption}
\usepackage{subcaption}
\usepackage{csquotes}
\usepackage{xcolor,soul}
\usepackage{graphicx}
\usepackage{wrapfig}
\usepackage{comment}

\newcommand\blfootnote[1]{%
  \begingroup
  \renewcommand\thefootnote{}\footnote{#1}%
  \addtocounter{footnote}{-1}%
  \endgroup
}

\title{Unpacking Large Language Models with Conceptual Consistency}

\author{Pritish Sahu$^{1,2\dagger*}$ \quad Michael Cogswell$^{1*}$ \quad Yunye Gong$^1$ \quad Ajay Divakaran$^1$ \\
$^1$SRI International\\
$^2$Rutgers University\\
}

\iclrfinalcopy 
\begin{document}

\maketitle
\blfootnote{$^\dagger$Work was done while interning at SRI International.}
\blfootnote{$^*$These authors contributed equally to this work.}
\blfootnote{ps851@cs.rutgers.edu, \{michael.cogswell,yunye.gong,ajay.divakaran\}@sri.com}

\begin{abstract}
If a Large Language Model (LLM) answers ``yes'' to the question ``Are mountains 
tall?'' then does it know what a mountain is?
Can you rely on it responding correctly or incorrectly to other questions about mountains?
The success of Large Language Models (LLMs) indicates they are increasingly able to
answer queries like these accurately, but that ability does not
necessarily imply a general understanding of concepts relevant to the anchor query.
We propose conceptual consistency to measure a LLM's understanding of relevant
concepts. This novel metric measures how well a model can be characterized by finding out how consistent its 
responses to queries about conceptually relevant background knowledge are.
To compute it we extract background knowledge by traversing paths
between concepts in a knowledge base and then try to predict the model's response to
the anchor query from the background knowledge.
We investigate the performance of current LLMs in a commonsense reasoning setting
using the CSQA dataset and the ConceptNet knowledge base.
While conceptual consistency, like other metrics, does increase with the scale of the LLM used, we find that popular models do not necessarily have high
conceptual consistency.
Our analysis also shows significant variation in conceptual consistency
across different kinds of relations, concepts, and prompts.
This serves as a step toward building models that humans can apply a theory of
mind to, and thus interact with intuitively.
\end{abstract}
\section{Introduction}
Large Language Models (LLMs) have had many exciting recent successes. These include
high performance and even emergent capabilities using just zero or few-shot prompting~\citep{gpt3,emergent_llm}, but
overall performance is still low compared to humans on a wide range of tasks for even the largest models~\citep{big_bench}.
A popular explanation of low performance and inconsistencies is that LLMs are simply learning to
mimic the data used to train them, and this basic pattern recognition limits generalizability,
in the case of LLMs exposing the limits of any understanding~\citep{paradox_reason_from_data,climbing_nlu}.
We would use a similar line of reasoning to guess whether a LLM will judge the following statement to be true or false:

\begin{displayquote}
The peak of a mountain almost always reaches above the tree line.
\end{displayquote}

If it performed well on examples from the same distribution we would say it
is likely to get it right or vice-versa if performed poorly on those examples.
Though valid, this explanation is incomplete because it is completely 
agnostic to the specific content of the statement.
We would apply the exact same reasoning and come to the same conclusion
for similar statements about say blood banks or mock trials, as long
as they were from the same distribution (in this example, the CSQA2 dataset~\citep{csqa2}).
This is in contrast to our day to day life, where our Theory of Mind 
allows us to understand other
agents (people) by attributing
beliefs, intentions, and desires to them~\citep{premack_woodruff} in a way that allows us to
usefully predict their behavior~\citep{machine_tom,dennett91}.
Beliefs are most relevant here, and should be conceptual in order to best
support human understanding~\citep{concept_explanations}.
Ideally we would also be able to apply this kind of understanding to 
LLMs, predicting that the model will be correct if it knows about mountains and tree lines or that it will be false if it has not
yet learned about tree lines.
This would be a conceptual model of the LLM that allows us
to predict its behavior.

We want to build models for which that kind of understanding is possible, so in 
this work we take a step toward that goal by modeling the conceptual
knowledge of a LLM and predicting when the LLM will be correct from that model.
While predicting performance based on training distribution falls cleanly out of ML theory,
our approach does not, so we test LLMs to see if they can be understood in this way.

\begin{figure}[!t]
    \vspace{-2.8em}
    \centering
    \includegraphics[width=0.8\textwidth]{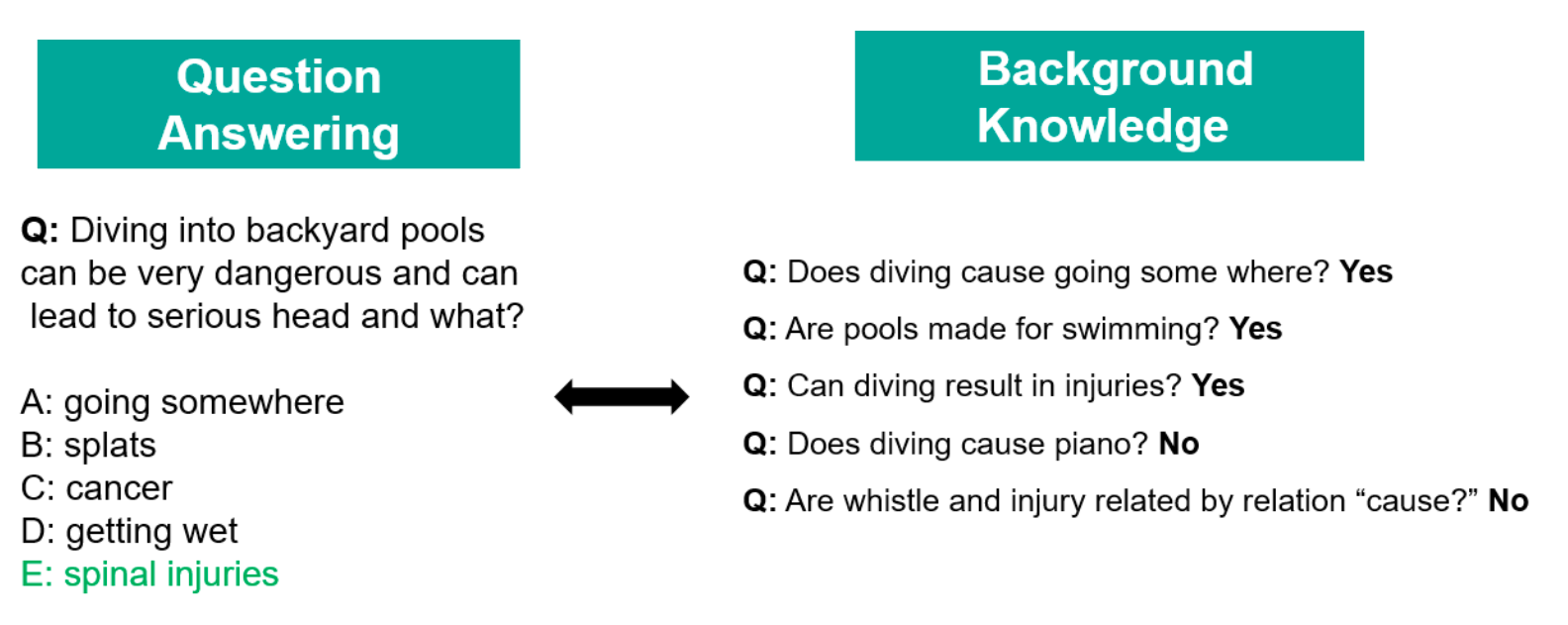}
    \caption{Example question and related background knowledge. A model is conceptually consistent when its knowledge of relevant background information is consistent with its ability to answer questions correctly.}
    \label{fig:teaser}
    \vspace{-1.5em}
\end{figure}

Our conceptual model is based on a direct examination of background knowledge relevant to a particular
anchor task (e.g., question answering) and a measurement of how well the background is known by the model.
From this and a measurement of question answering performance
we compute \emph{conceptual consistency} (\autoref{fig:teaser}),
quantifying whether a model's knowledge of relevant background is consistent
with its ability to perform the anchor task.

Background knowledge is any knowledge relevant to a particular example. We distinguish between
constructive relevance and conceptual relevance. Background facts are constructively relevant if they can be used to logically support the correct conclusion. One might use constructive background knowledge to reason that mountain peaks are typically above the tree
line because tree lines typically fall far short of mountain peaks and
trees are not nearly as tall as mountains. That ``tree lines typically fall
far short of mountain peaks'' and ``trees are not nearly as tall as mountains''
would be constructive background knowledge. On the other hand, conceptual background knowledge
need only share relevant concepts, so ``trees grow on mountains'', ``mountains can have tundra'',
and  ``trees do not grow on tundra'' are all examples of conceptually relevant background knowledge.
In this paper, we focus on conceptual relevance.

After extracting background knowledge we use prompting to measure how
a given LLM handles the background knowledge and
how it performs at the anchor task.
For this we study three varieties of generative
language models across multiple scales up to 66 billion
parameters and use a majority vote style prompting procedure
to maximize the robustness of our approach to linguistic
variations.

Our core contributions are
\begin{itemize}
\item
We extract conceptually relevant background knowledge with respect to examples from an anchor task and map them onto background knowledge questions.

\item
We use a novel majority vote style zero-shot prompting procedure applied to generative language models to measure LLM performance.

\item
We measure conceptual consistency, focusing on generative language models and showing that consistency is low though it does increase with model size up to 66 billion parameters.

\item
We report conceptual patterns in model behavior that fall out of our analysis.

\end{itemize}

\section{Related Works}

Much work has been devoted to studying the limits of large language models beginning with BERT~\cite{bert}.
Typical evaluation of language models measures performance on datasets of labeled examples
across a range of tasks, such as those that constitute the
amalgamated BIG Bench benchmark~\citep{big_bench}.
In general they are explained as simply mimicking the data they were trained with.
Low level critiques question the ability of LLMs to
reason logically~\citep{paradox_reason_from_data} or pass adapted psychological tests of
language in humans~\citep{what_bert_is_not}.
High level critiques question the ability of LLMs to understand anything at all~\citep{climbing_nlu}, though
one alternative avoids this by defining meaning based on conceptual role~\citep{meaning_without_reference}.

\paragraph{Consistency}
Most relevant here is the large literature that studies failures in consistency
of these models.
A common approach is to verify an expectation about how the outputs of two different prompts should be related. For example, the prompts ``Is the rose red?'' and ``What color is the rose?'' 
with the expectation that answering ``no'' to the first question is inconsistent with answering ``red'' to the second~\citep{are_roses_red}.
CheckList~\citep{checklist} offers a taxonomy that can be applied to these approaches.
Invariance based consistency considers two versions of the input (often paraphrases) where
the output is expected to be the same
for both versions ~\citep{negated_misprimed,pararel,convqa,calum,checklist}.
Directional expectation consistency expects a specific change in the output based on how the inputs
are constructed~\citep{are_roses_red,polyjuice,negated_misprimed,calum,checklist} --
e.g., that one output should be the negation of the other.
These works generally find that the models they study are often inconsistent, with varying
degrees for various kinds of consistency.
Many also improve their consistency measure by fine-tuning the studied models  for it.

These kinds of consistencies are constructive rather than conceptual.
They establish and improve the degree to which language
models can be said to have beliefs~\citep{lm_beliefs}, but in order for beliefs to support
a theory of mind they should be actionable in that they should support predictions about
another agent's behavior~\citep{machine_tom,dennett91}.
That a LLM has logically consistent outputs does not imply it will be 
right, but a theory of mind can provide evidence for or against correctness.
This looser definition of consistency also makes the idea more general.
In contrast, previous work,  especially for more complex definitions of consistency,
has relied on domain specific and expensive annotated datasets
to  measure consistency.

Furthermore, in this work we focus on large generative language models as opposed
to the masked language models where most prior work on consistency is focused.
Some of the greatest success has come from scaling generative LLMs~\citep{gopher_scaling,emergent_llm},
however most existing work on consistency focuses on smaller scale models
(mostly masked style language models)
because large scale generative models were not openly available until
recently~\citep{opt}, and also because fine-tuning the larger models
is prohibitively expensive, making it infeasible as a way to improve consistency.
It may be that even though performance increases with scale consistency does not, so in
this work we study model scale to the degree possible with open source models and a
limited compute budget.

\paragraph{Explanations and Interpretability}
Explaining neural net decisions has been studied for a long time, but recent
work has shifted focus towards concept based explanations~\citep{concept_explanations}.
For example, derivatives can be used to quantify conceptual influence
in neural nets~\citep{tcav} and model the causal importance of concepts~\citep{cace}.
Another line of work links theory of mind to explanations~\citep{theory_of_ais_mind,tom_explanation}.
We use concepts to explain LLMs from a theory of mind perspective.
    
\section{Proposed Method}
To measure conceptual consistency we need to measure background knowledge and QA performance then predict the latter from the former.
First we describe extraction of background knowledge in the form of questions
with known answers,
second we describe how we use prompting to answer both background and anchor questions,
and finally we describe the conceptual consistency metric which correlates the
two by predicting one from the other.

\subsection{Background Knowledge Extraction}\label{sec:bk_extract}
Here we focus on question answering (QA) problems and assume a knowledge base of
content relevant to the specific QA task is available. Examples in our QA dataset
consist of a question $Q$ with corresponding choices $S = \{s_1, \ldots, s_{|S|}\}$,
one of which is the correct answer $A$. These choices can be single words or short
phrases.
We call $(Q, S, A)$ the anchor query because our first task is to find conceptually relevant background knowledge with respect to that information.

The background knowledge for a given anchor corresponds to a list of facts
in a knowledge base. We assume each fact $F = (c^1, r, c^2)$ is represented by two
concepts $c^1$ and $c^2$, and a relation $r$ between those concepts.
Our task is to extract a set $B = \{f_1, \ldots, f_{|B|}\}$ of facts conceptually
relevant to the anchor and then map those facts onto questions that can be asked
to roughly determine whether the LLM knows the background knowledge.

\paragraph{Extracting Background Facts}
We conceive of the knowledge base as a graph that connects concepts as nodes and
relations as edges. The set of concepts $C$ in the anchor query is 
all the short meaningful words and phrases that appear in any part of the anchor 
query and have overlap ($> 50\%$ of words match) with a concept from the knowledge base.
For two different concepts $c^1, c^2 \in C$ we consider all tuples from all
paths which connect those concepts in the knowledge base, forming a cloud of
relational information which constitutes the background knowledge for the
anchor given by the selected knowledge base. These are conceptually relevant
because each tuple either shares a concept with the anchor or is somehow
connected to such a concept via some relation.
This is broader than the set of tuples that logically support the correct answer or refute incorrect answers.

In practice this list of background
knowledge tuples is too large, so we need to restrict it to a more manageable
yet still relevant list. We do this by setting the maximum path length to 1,
essentially looking for concepts which appear in the anchor and are directly
related to each other in the knowledge base. This case follows \cite{towards_gn},
where they were interested in extracting knowledge tuples to be inserted 
into neuro-symbolic models.

\paragraph{Background Questions}
In order to measure how well a LLM already knows the content captured by
these tuples we automatically translate them into natural language questions.
Consider a fact tuple $(c^1, r, c^2)$. We substitute its three variables into
a natural language template designed for the specific relation $r$. For example, the tuple (\texttt{book}, \texttt{used for}, \texttt{school}) would be translated into ``Are book used for school?''
Note that because the tuple exists in the knowledge base we know the correct answer
to be some version of ``yes''.
The templates we use for each relation are included in \autoref{tab:prompt_style} of the appendix.

\paragraph{Negative Background Knowledge}
So far we have only background tuples where the relation $r$ really does exist
between $c^1$ and $c^2$, so the correct answer is always ``yes.''
Language models are often biased towards certain outputs
and in this case we found a ``yes'' bias to be particularly strong in some models,
especially the smaller versions of OPT (\autoref{fig:opt_bias}).
As a result those smaller models can outperform the larger ones even when they understand
the background knowledge less, just because they are prone to saying ``yes'' to everything.
We fix this by extracting negative background knowledge tuples -- to which the correct
answer is some version of ``no'' -- to mirror the positive ones.

\begin{figure}[t]
    \vspace{-2.8em}
    \centering
    \includegraphics[width=0.89\textwidth]{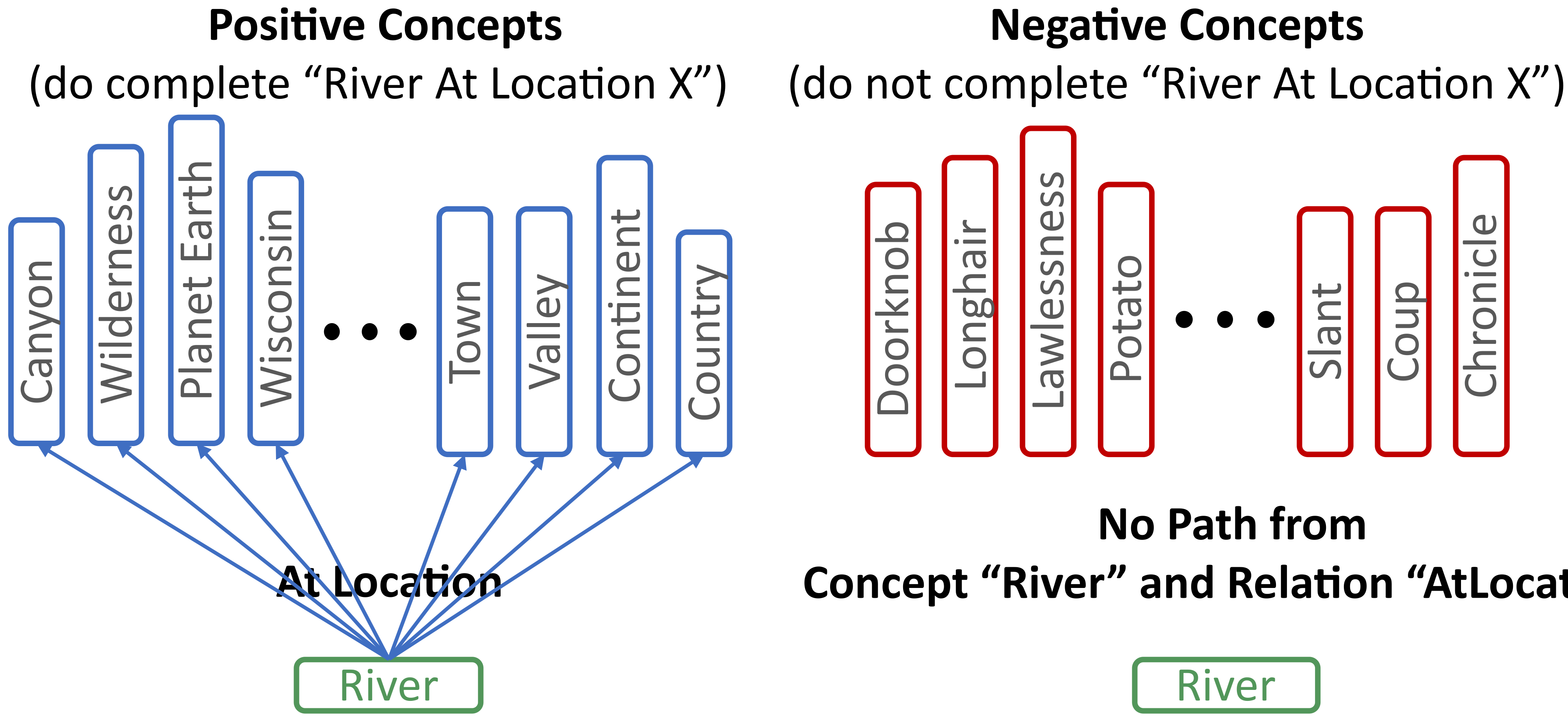}
    \caption{Facts from knowledge bases are true by virtue of being in the
    knowledge base. We mine negative background facts by finding concept pairs
    without an edge to ensure our extracted
    background knowledge is balanced.}
    \label{fig:neg_tuples}
    \vspace{-1.0em}
\end{figure}

We frame the problem in a pairwise fashion: given a positive tuple $(c^1, r, c^2)$
the negative tuple generation task is to find an alternative $\bar{c}^2$ for which
the correct answer is ``no.''
The pairwise nature ensures that we measure background knowledge in a balanced fashion
to best address the ``yes'' (or ``no'') bias issue.
As depicted in \autoref{fig:neg_tuples}, we form a set of
choices $\bar{C}^2$ that includes every concept $\bar{c}$ in the knowledge
base that meets 3 criteria:
\begin{enumerate}
\item $\bar{c}$ does not form a valid tuple $(c^1, r, \bar{c})$,
\item $\bar{c}$ is not cyclic (i.e., not equal to to $c^1$), and
\item $\bar{c}$ is in the English dictionary.
\end{enumerate}
Our final choice for $\bar{c}^2$ is a single sample from the
uniform distribution over $\bar{C}^2$. The amount of concepts in the English dictionary is quite large, hence we curate the distribution space of $\bar{C}^2$ after applying criteria 1 and 2 to keep most frequently used concepts.
We make this choice in advance, treating background tuples as a dataset, so that
even if a positive background tuple is used for multiple anchor queries it is always paired
with the same negative tuple.
However, if there are multiple choices of $c^2$ for the same $c^1$ and $r$ then
we sample $\bar{c}^2$ independently for each of those choices.
The final set of background tuples for an anchor query includes all positive tuples
that match that anchor along with all negative tuples that are paired with the positives.
Examples of extracted negative background facts can be found in the appendix at \autoref{tab:neg_tuples}.

\subsection{Answering Background and Anchor Questions with Prompting}
Now we need to actually extract an answer to a given background question or anchor
question from the model.
We could simply provide the resulting text as input to the language model and then
see what most likely word it generates next. If it generated the word ``yes'' we would say
the LM knew this background fact and otherwise we would say it did not. However, in
experiments mainly with the OPT models we found this process to be highly noisy,
with inconsistent performance across models and relations.
This is probably due to inconsistencies in the models themselves~\citep{pararel}, but
also the simple nature of our templating procedure which doesn't always get the
grammar correct.
We are interested in measuring the language model's
\emph{conceptual} consistency, not \emph{linguistic} consistency, so we implemented a
majority-vote-style procedure to make our prompting procedure more robust to 
variations in the specific tuple-to-prompt mappings.

Instead of using just one question and one answer we consider many variations on the same
question and many potential answers, taking the combination assigned the highest
likelihood by the model as its predicted answer to the question.
This approach constrains the output space of the model by limiting the words
over which likelihood is maximized.

To vary answer language we chose a
list of positive and negative words as potential answers including $\{$(Yes, No), (True, False), (Right, Wrong), (Correct, Incorrect), (Positive, Negative), (Pass, Fail), (On, Off)$\}$.
We then chose the one with the highest likelihood as our model's answer.
To vary question language we substitute the question generated from our templates
into \emph{meta-prompts}, which are minor variations on how to
ask the question. For example, the previous question might also be phrased as
``Answer this question as 'Yes' or 'No'. Question: Are books used for school?''
The meta-prompts we used, which can also take pairs of possible labels 
are reported in \autoref{tab:metaprompts}.

We found this variation to be essential for achieving some consistency across linguistic
variations when experimenting with the OPT language models~\citep{opt}.
Voting over multiple variations is similar to the strategies adopted in a number of recent approaches.
In \citet{vered_unsup} clarifying statements and answers to questions are chosen
by taking a maximum over likelihood scores of different possibilities, but they do not
use this to account for linguistic variation in prompts.
Similarly, in \citet{pararel} knowledge base tuples are completed by picking a choice
from a restricted candidate set; here it is not used to control linguistic variation
in order to study another variable, but is used to study linguistic variation itself.
More recently majority vote has been used to evaluate generative models when
generating longer snippets of text~\citep{self_chain_of_thought,reason_chain_of_thought}.
These works use the approach to account for linguistic variation, but in the model
outputs (more than just a few words) rather than in the inputs (i.e., the prompts).
In comparison to all those studies, our majority-vote-style procedure is used to account for linguistic
variation in the prompts themselves.

\begin{table}[t]
\vspace{-2.8em}
\centering
\caption{Meta-prompts utilized for question generation.}
\resizebox{0.99\textwidth}{!}{%
\begin{tabular}{l}
\hline \noalign{\smallskip}
\textbf{Meta-Prompts} \\
\hline \noalign{\smallskip}
1. \texttt{<question>}? \\[3pt]
2. \texttt{<question>}. Is this true? \\[3pt]
3. Answer this question as '\texttt{<label\_a>}' or '\texttt{<label\_b>}'. Question: \texttt{<question>}? \\[3pt]
4. Each item is a question and answer. Answer is one of '\texttt{<label\_a>}' or '\texttt{<label\_b>}'. Question: \texttt{<question>}? Answer: \\[3pt]
5. Pick '\texttt{<label\_a>}' or '\texttt{<label\_b>}'. Question: \texttt{<question>}?Answer: \\
\hline \noalign{\smallskip}
\end{tabular}
\label{tab:metaprompts}
}
\vspace{-1.0em}
\end{table}

\subsection{Measuring Conceptual Consistency}

Now we are almost ready to measure conceptual consistency.
Till now we have extracted answers $\hat{A}_B^{i,b}$ for the $b$th background knowledge
question $Q_B^{i,b}$ for the $i$th example in the anchor task dataset.
The anchor questions and answers are $Q_A^i$ and $\hat{A}_A^i$.
We translate these questions and answers 
into scores that measure how well the model knows the background knowledge
and how well it performs at the anchor task.
These background and task scores are defined respectively \emph{for each anchor example} using accuracy
\begin{align}
S_B^i &= \left(
    \E_{b \in \mathcal{P}_i}
        \left[ \ind{A_B^{i,b} == \hat{A}_B^{i,b}} \right] +
    \E_{b \in \mathcal{N}_i}
        \left[ \ind{A_B^{i,b} == \hat{A}_B^{i,b}} \right]
\right) / 2 \label{eqn:bkg_score} \\
S_A^i &= \ind{A_A^i == \hat{A}_A^i  } \label{eqn:anchor_score}
\end{align}
where $A_B$ and $A_A$ are the correct answers to those questions,
$\mathcal{N}_i$ is the set of negative background tuples for anchor example $i$,
$\mathcal{P}_i$ is the set of positive background tuples for anchor example $i$,
and $\ind{.}$ is the indicator function.
Note that the background score weights negative and positive tuples evenly.

Finally we compute the conceptual consistency of a model on a given dataset and knowledge
base by predicting the task score from the background score and reporting average precision.
At a given threshold $t \in [0, 1]$ our predicted task score is
$\hat{S}_A^i = \ind{S_B^i >= t}$.
This score predicts the model will be correct when it is 1 or 
incorrect when it is 0.
Intuitively, this score will be high when the model answered more background knowledge
questions correctly, so we are predicting that the model will perform better at this
Now conceptual consistency is
\begin{align}
CC &= AP(S_A^i, \hat{S}_A^i) \label{eqn:cc}
\end{align}
where $AP(\cdot)$ measures the average precision of the $\hat{S}_A$ predictor. The positive category is the one where the model is correct.

\section{Experiment Setup}
\paragraph{Dataset and Knowledge Base}
We conduct zero-shot evaluation on CommonsenseQA (CSQA)~\citep{talmor2019commonsenseqa} task and use ConceptNet~\citep{conceptnet} as our knowledge base.
CommonsenseQA covers a broad range of common sense concepts, with each entry containing a question and 5 answer choices.
Crucially CommonsenseQA is derived from the ConceptNet graph, so we expect the knowledge base to have relevant background information.
We use a subset of 14 out of the 36 available relations most relevant to common sense, including ``antonym", ``at location", ``capable of", ``causes", ``desires", ``form of", ``has a", ``is a", ``made of", ``part of", ``related to", ``similar to", ``synonym", and ``used for".
Our experiments used the development set, since test set answers are not publicly available. No model training or hyperparameter tuning was applied to our pre-trained LLMs.
Most of our development effort went into creating stable prompts.

\paragraph{Models}
We probe for conceptual consistency using three publicly available model families, including OPT-$\{350M,1.3B,13B,30B,66B\}$~\citep{opt}, GPT(EleutherAI)-$\{125M,2.7B,6B\}$~\citep{gpt_neox} and T0-$\{3B,11B\}$~\citep{t0}. The OPT and GPT(EleutherAI) models were chosen
because they are some of the largest publicly available models and have a range of checkpoints over sizes.
We use T0 because it was tuned specifically for zero-shot prompting setting and achieved competitive results in zero-shot generalization.
The prompts and general approach remain the same for all models, though
we did most of the development using OPT models.

\section{Results}
Here we report the conceptual consistency of the models we study,
analyze the individual background and task components of that score,
show qualitatively how performance on relations and concepts varies,
and analyze bias related to our prompting approach.

\begin{wrapfigure}{r}{0.55\textwidth}
    \vspace{-1.0em}
    \centering
    \includegraphics[width=0.5\textwidth]{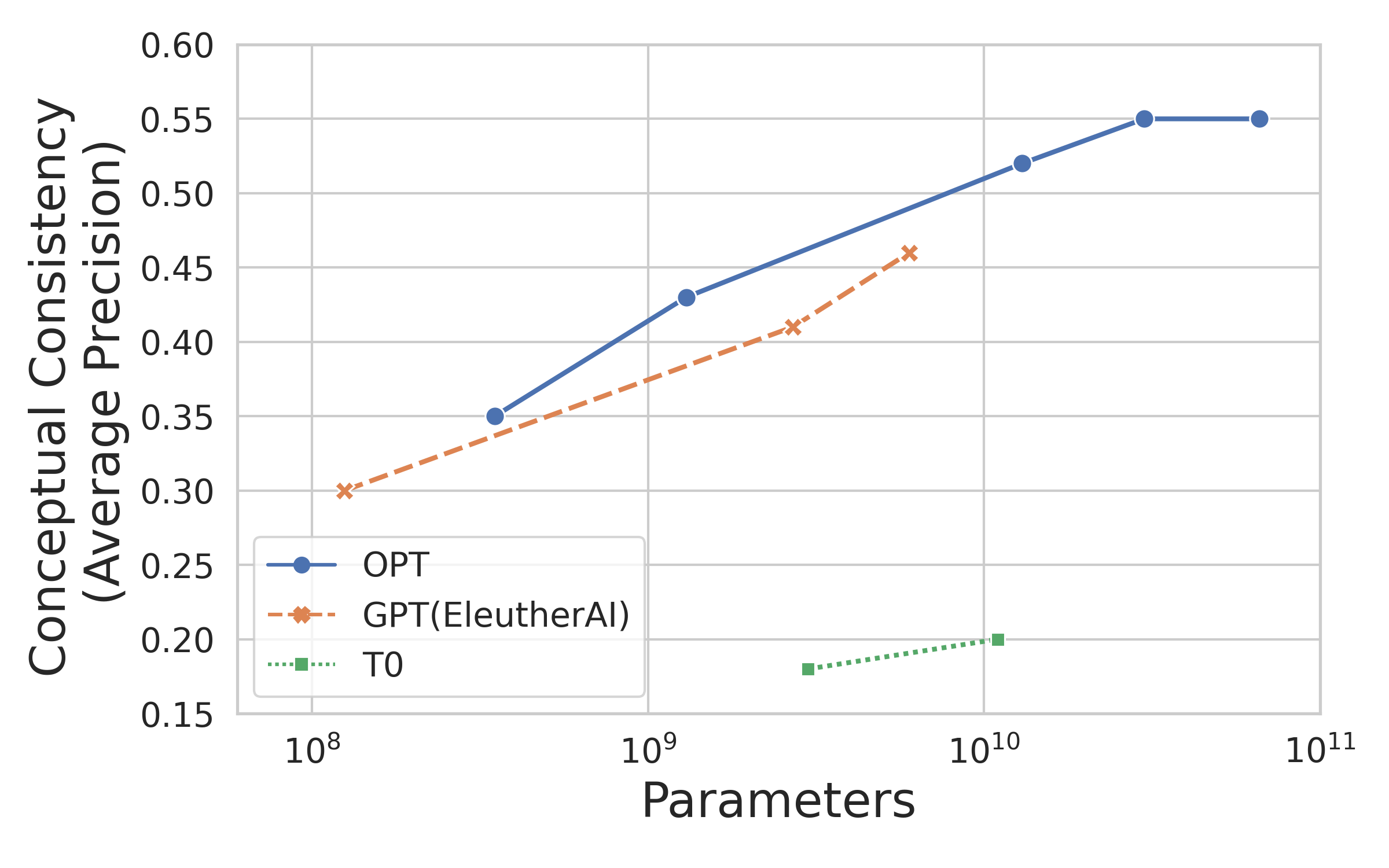}
    \caption{Conceptual Consistency of Language Models. This measures
    our ability to predict whether a language model will be correct
    from its knowledge of relevant background information.}
    \label{fig:ap_score}
    \vspace{-1.5em}
\end{wrapfigure}

\paragraph{Conceptual Consistency}
We compute the conceptual consistency (\autoref{eqn:cc}) for each LLM
and report the results in \autoref{fig:ap_score}.
In absolute terms average precision is on the lower end, showing
significant conceptual inconsistency.
A model's knowledge of background information is somewhat predictive
of its question answering correctness.
Our results show that conceptual consistency generally grows with the scale of the model across all model classes,
indicating that bigger models are not just more accurate, but also more consistent.
A notable exception is between OPT-30B and OPT-66B, where there is no improvement from increased size.
Both achieve the highest consistency among all models, but this one
data point suggests there may be diminishing returns.
OPT models perform best, with GPT models close behind and T0 demonstrating much worse performance.
This indicates that OPT and GPT are much more conceptually consistent than T0.
The difference could be due to the decoder only nature of OPT and GPT.
It could also reflect a degree of forgetting in T0 due to supervised
fine-tuning, but this is doubtful because one of the many datasets
T0 was trained on was CSQA.
However, it may also reflect the fact that our prompts are
not designed specifically for T0.

\begin{figure}[hb]
    \vspace{-0.5em}
    \centering
    \begin{subfigure}[t]{0.49\columnwidth}
        \includegraphics[width=\columnwidth]{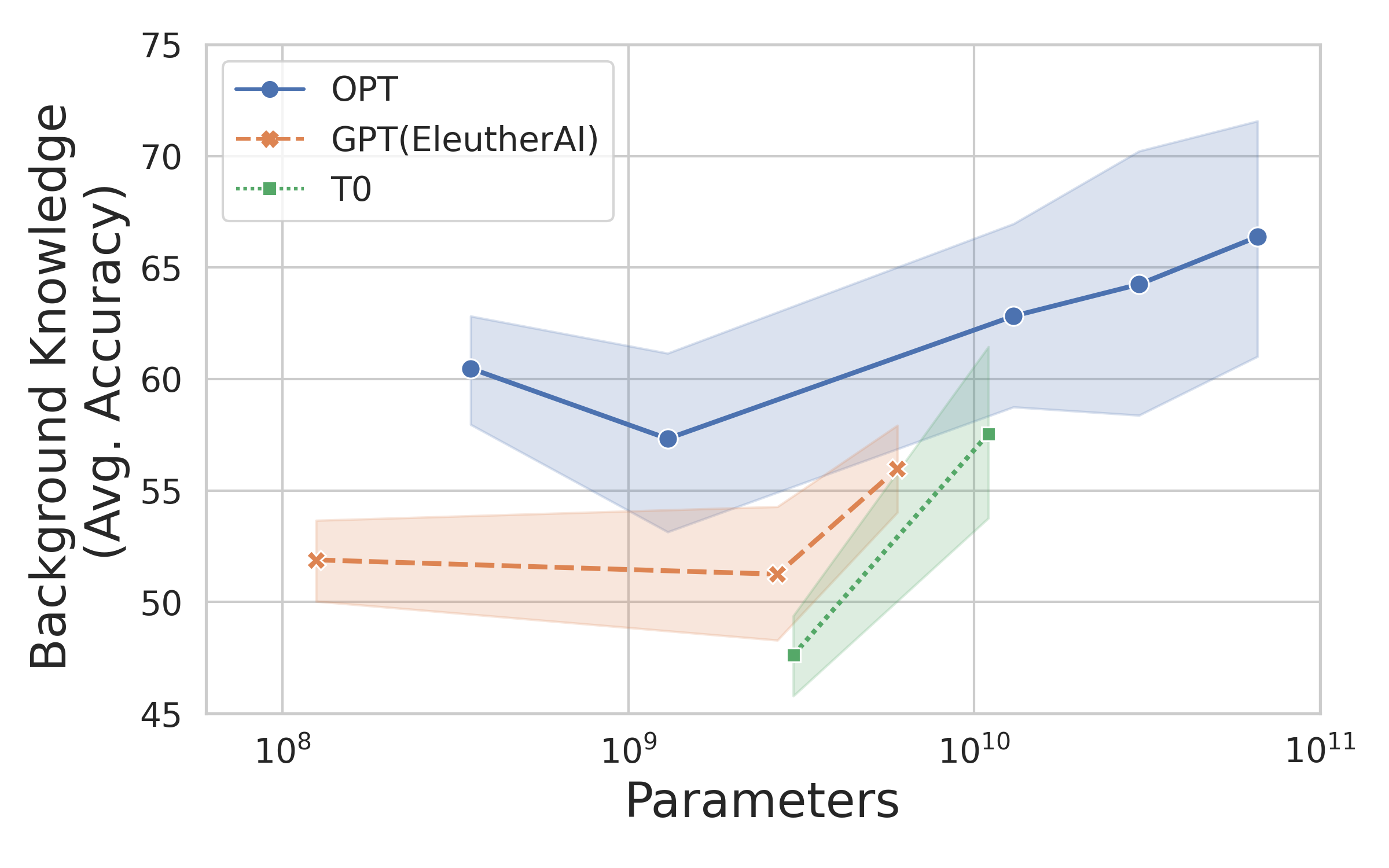}
        \caption{
            Background knowledge performance: how good language models are
            at verifying whether our extracted background facts are true/false.
            Averaged over 14 relations.
        }
        \label{fig:bkg_knowledge}
    \end{subfigure}
    \hfill
    \begin{subfigure}[t]{0.49\columnwidth}
        \includegraphics[width=\columnwidth]{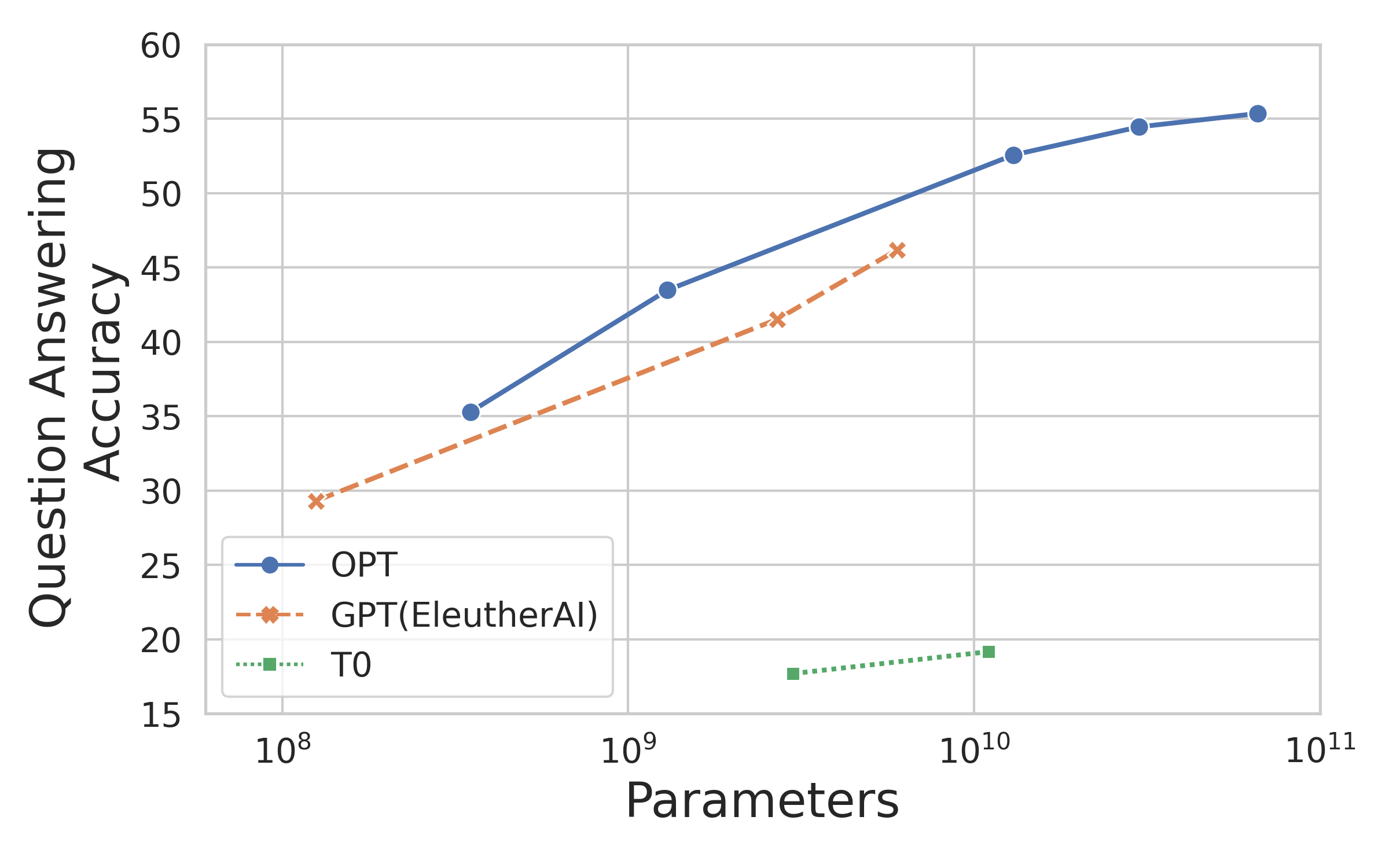}
        \caption{Anchor Task (CommonsenseQA question answering) performance measured
        by our zero-shot prompting approach.}
        \label{fig:task_knowledge}
    \end{subfigure}
    \caption{Aggregated background performance and anchor task performance.}
    \label{fig:indep_knowledge}
    \vspace{-1.0em}
\end{figure}

\paragraph{Background and Task Performance}
We also measure the components of conceptual consistency in
\autoref{fig:indep_knowledge}.
For background knowledge we compute \autoref{eqn:bkg_score} averaged per relation
and then averaged over all 14 relations.
This is reported with a 95\% confidence interval in \autoref{fig:bkg_knowledge}.
There is an imbalance in how often these relations occur in the data,
so this approach weights relations equally just for the purpose of
assesing background knowledge.
For task knowledge we report the anchor score (\autoref{eqn:anchor_score}) in \autoref{fig:task_knowledge}, which is CSQA accuracy in these experiments.
In both cases our prompting approach is able to extract correct answers
from our LLMs.
The trend in both cases is again an increase in performance with model
size. This was expected in both cases, but it is interesting to note that the range of performance is smaller for background knowledge,
suggesting that increasing size helps task knowledge more than background knowledge.
Intuitively question answering is a more complex skill that builds on 
the less complex background knowledge completion skill. 
From that perspective these results are also an evidence of
a skill hierarchy like Bloom's Taxonomy~\citep{sahu_bloom},
where simpler skills are learned before more complex ones.
As for conceptual consistency, OPT models have higher performance compared to GPT and T0 models, and this is likely due to the same
reasons discussed previously.
It is however notable that the gap between T0 and the other models
is much smaller for background knowledge.
Also, we see a marginal increase in performance between OPT-30B and OPT-66B
for both task and background knowledge, though it may not be significant.

\paragraph{Background vs Consistency}
\begin{wrapfigure}{r}{0.68\textwidth}
    \vspace{-1.0em}
    \centering
    \includegraphics[width=0.67\textwidth]{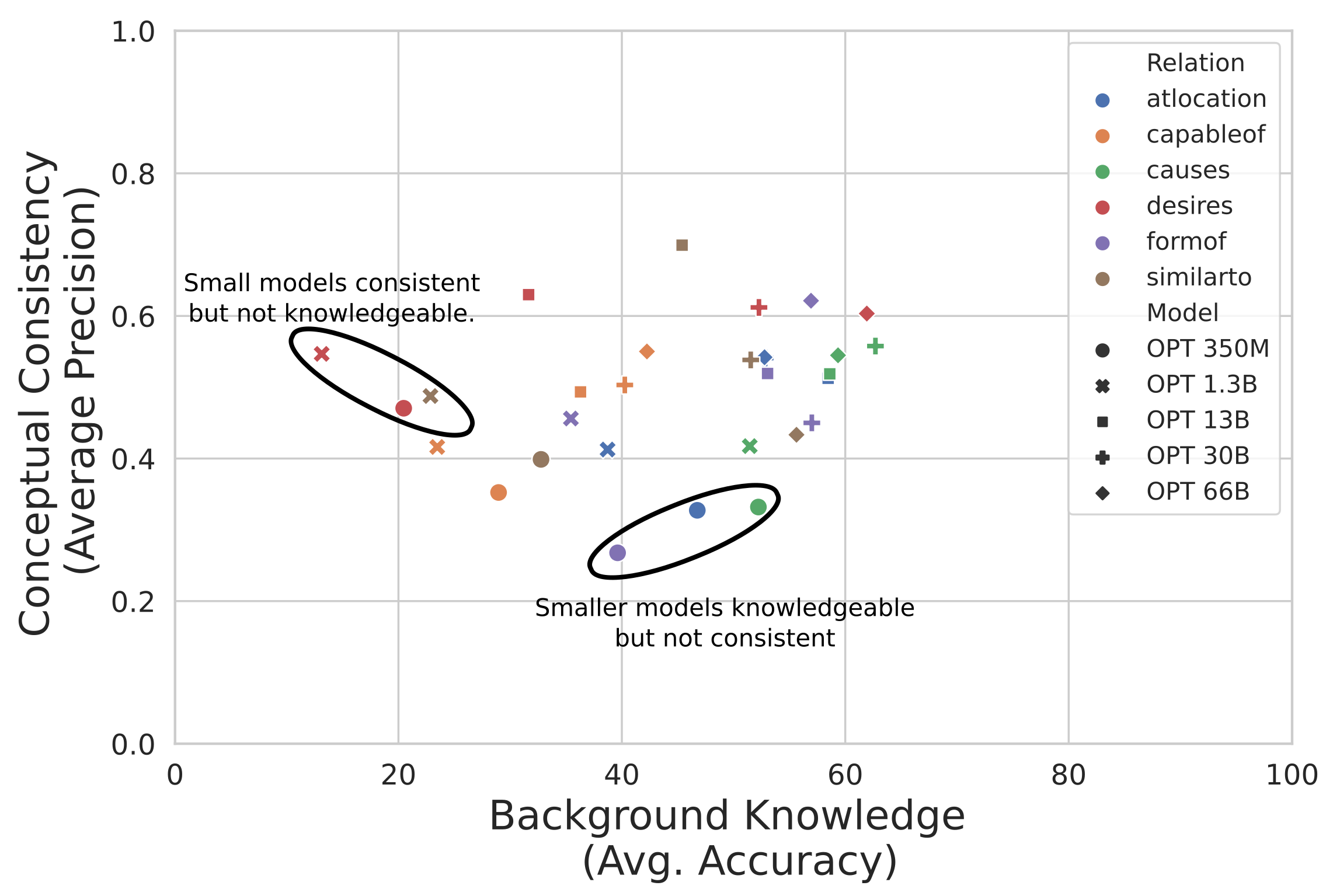}
    \caption{Scatter plot of Background Knowledge vs Conceptual Consistency.}
    \label{fig:bk_vs_ap_score}
    \vspace{-1.5em}
\end{wrapfigure}
Next we ask where background knowledge and consistency diverge.
In \autoref{fig:bk_vs_ap_score} we plot background score versus conceptual
consistency for 6 relations that seemed to be more like outliers.
Smaller models tend to be further from the diagonal, either being
inconsistent but knowledgeable or consistent without knowing much.
Relations also behave differently. Small models don't know the
``desires" relation very well, but they are somewhat conceptually
consistent about it, to the point that even though large models get
to know the relation better they do not get more consistent about it.
In the reverse direction, all model scales know roughly the same
amount about ``causes" background information, but the
larger models are much more conceptually consistent about it.

\begin{figure}[t]
    \vspace{-2.8em}
    \centering
    \begin{subfigure}[t]{0.49\textwidth}
        \includegraphics[width=\textwidth]{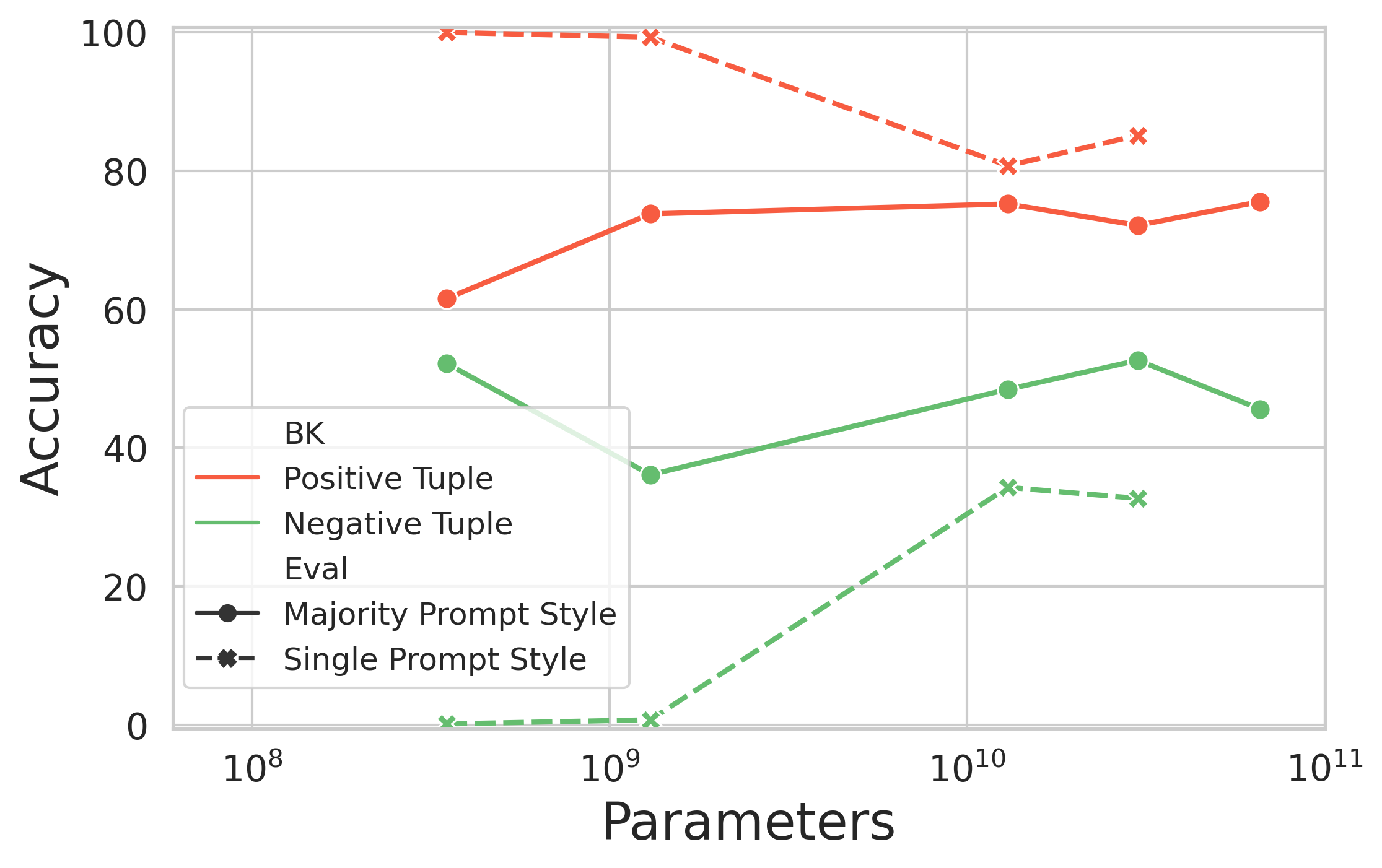}
        \caption{OPT models with our majority style prompting and with a
        single prompt.}
        \label{fig:opt_bias}
    \end{subfigure}
    \hfill
    \begin{subfigure}[t]{0.49\textwidth}
        \includegraphics[width=\textwidth]{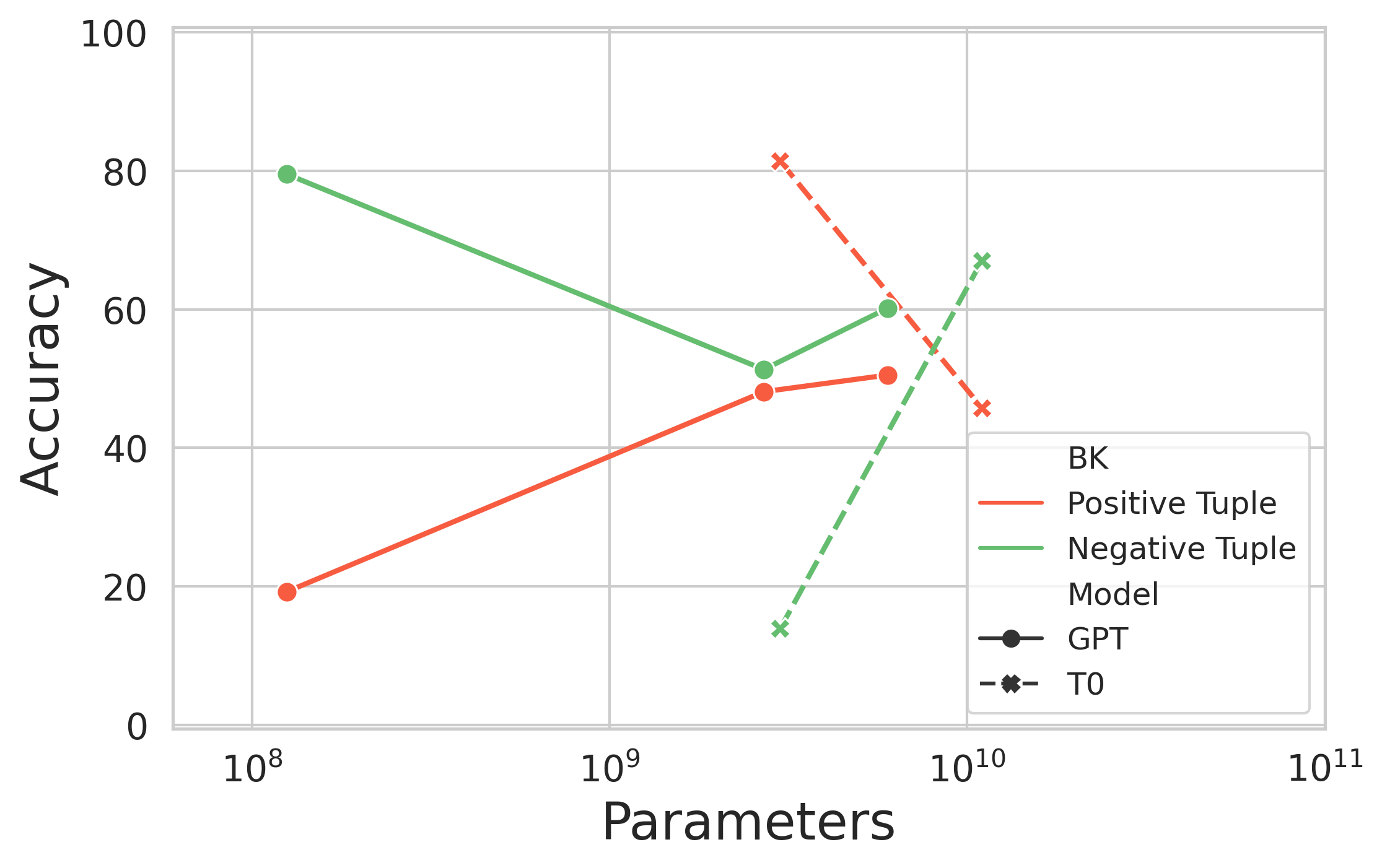}
        \caption{GPT and T0 Bias}
        \label{fig:gpt_t0_bias}
    \end{subfigure}
    \caption{Background knowledge performance for negative and positive facts.}
    \label{fig:bias}
    \vspace{-1.5em}
\end{figure}

\paragraph{Prompting Bias}
In \autoref{fig:bias} we analyze our majority-vote-style prompting
and prompting biases we found in our LLMs.
Each figure reports the background score (\autoref{eqn:bkg_score})
restricted to either the set of negative background facts or positive 
background facts, so accuracy reduces to the percentage of times
the model said a version of ``no" or ``yes," respectively.
\autoref{fig:opt_bias} reports this metric for the majority-vote-style prompting
we used (Majority Prompt Style) and a previous version where we tried
to find a single best meta-prompt and answer pair (Single Prompt Style).
In our single prompt experiments we found that the two smaller OPT 
models had a strong ``yes" bias as shown by the red dashed line
in \autoref{fig:opt_bias}. We introduced negative background knowledge,
which detected this problem (green dashed line), but we also found
that our majority-vote-style prompting helped ameliorate the issue
(solid lines).
Even with majority-vote-style prompting both T0 and GPT(EleutherAI)
still display a significant ``yes" (T0) and ``no" (GPT) bias
at smaller scales.

\begin{figure}[!t]
    \vspace{-2.8em}
    \centering
    \begin{subfigure}[]{\textwidth}
        \includegraphics[width=\textwidth]{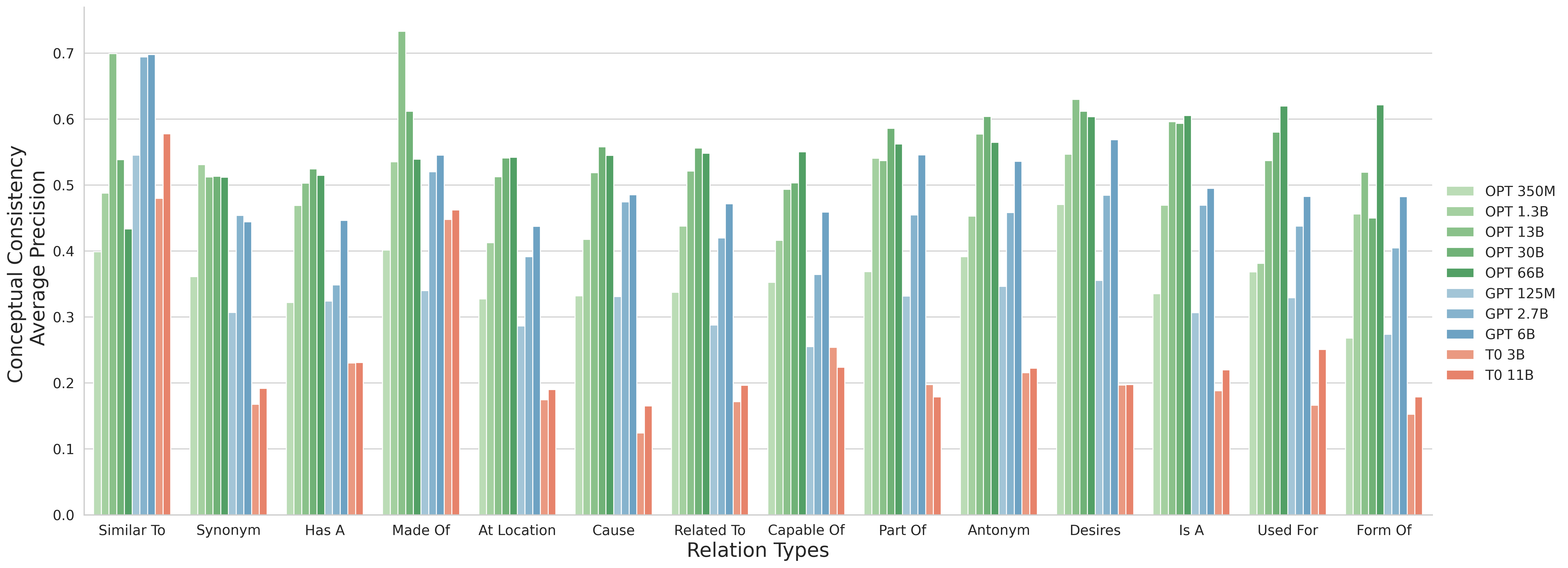}
        \caption{By Relation}
        \label{fig:per_relation}
    \end{subfigure}
    \begin{subfigure}[]{\textwidth}
        \includegraphics[width=\textwidth]{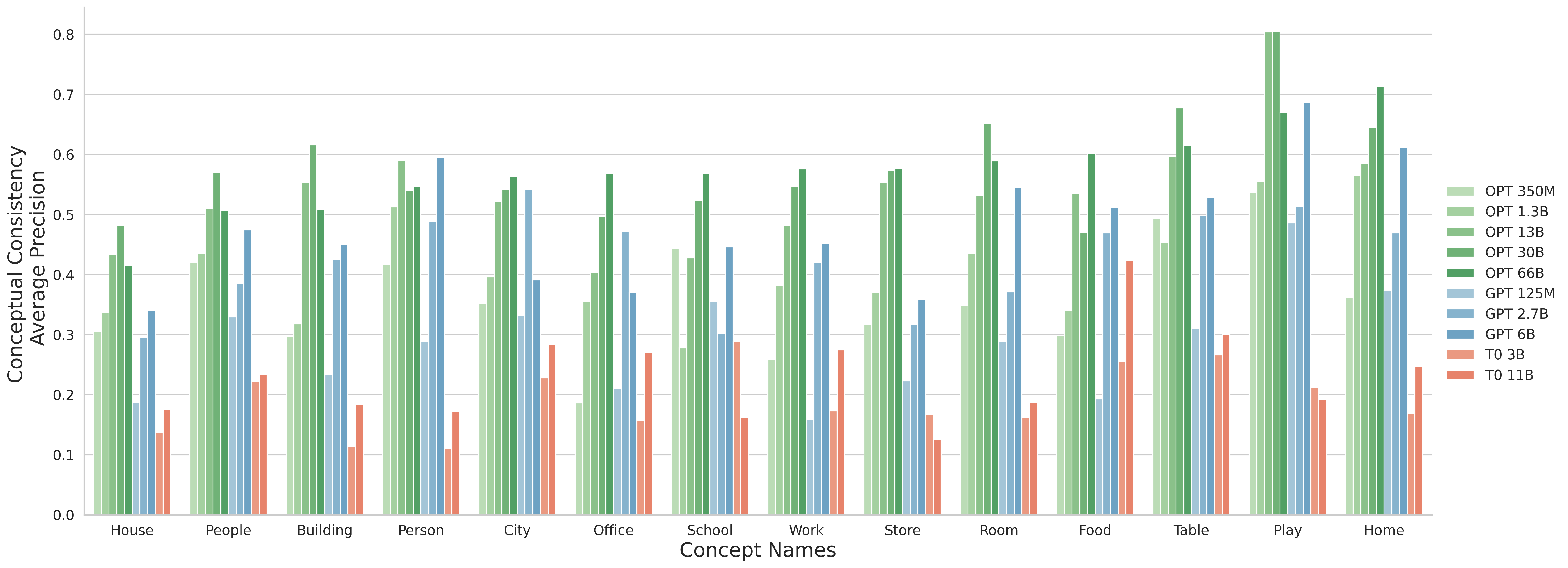}
        \caption{By Concept}
        \label{fig:per_concept}
    \end{subfigure}    
    \caption{
        Conceptual consistency computed for different subsets of CSQA examples
        partitioned according to whether the background facts for an example
        contain the relation or concept.
    }
    \label{fig:per_relation_concept}
    \vspace{-1.5em}
\end{figure}

\paragraph{Concept and Relation Level Consistency}
We analyze the conceptual consistency at the level of relations and concepts in \autoref{fig:per_relation_concept}. 
\autoref{fig:per_relation} shows consistency for all relations. 
In case of \autoref{fig:per_concept}, we picked 14 most occurring concepts with minimum occurrence count of 28.
Though the overall trend is that larger models know more background
knowledge, there are variations in which concepts and relations
a model is consistent about.
In some cases you don't need a larger model, or at least
larger models can have sub-par performance. For example,
our largest models (OPT-66B and OPT-30B) are outperformed by smaller
versions of OPT, GPT(EleutherAI), and even T0 on the ``Similar To"
relation. Less extreme differences occur for other relations like ``Made Of", and again
lesser differences occur for concepts like ``Play". This sensitivity of relations
indicates our prompts, which we design per relation, could be a factor.
Another observation is that the increase in performance with size 
is robust for GPT and T0, but that trend is more often violated
for the OPT models.
In general, trends in concept consistency are more robust than trends
in relation consistency.

\section{Conclusions}
In this paper we built a kind of theory of mind about LLMs to study their conceptual consistency, whether their knowledge
of relevant background information is consistent with their ability to
answer questions correctly.
For a question we extracted background knowledge from a knowledge base
of related concepts and used prompting to measure whether popular 
open source LLMs knew that background information.
We also measured their ability to answer common sense questions correctly.
This set us up to measure conceptual consistency by
predicting correctness from our background knowledge measure.
We found that LLMs have a moderate amount of conceptual consistency,
and that it increases with scale.
We also found that while knowledge of background information
increases with model scale it does not increase nearly as much as
correctness or conceptual consistency, indicating that models size
has a larger impact on difficult tasks than simpler ones and providing
evidence of a hierarchy of related skills.
In the future we would like to study whether this measure of consistency
can be used to guide how humans understand LLMs. We also want to focus further
on the hierarchy of skills by creating a dataset that tests multiple levels
of comprehension.

\subsubsection*{Author Contributions}

\subsubsection*{Acknowledgments}
We thank Karan Sikka for helpful discussions about using large language models.

\bibliography{iclr2023_conference}

\begin{thebibliography}{36}
\providecommand{\natexlab}[1]{#1}
\providecommand{\url}[1]{\texttt{#1}}
\expandafter\ifx\csname urlstyle\endcsname\relax
  \providecommand{\doi}[1]{doi: #1}\else
  \providecommand{\doi}{doi: \begingroup \urlstyle{rm}\Url}\fi

\bibitem[Bender \& Koller(2020)Bender and Koller]{climbing_nlu}
Emily~M. Bender and Alexander Koller.
\newblock Climbing towards nlu: On meaning, form, and understanding in the age
  of data.
\newblock In \emph{ACL}, 2020.

\bibitem[Black et~al.(2022)Black, Biderman, Hallahan, Anthony, Gao, Golding,
  He, Leahy, McDonell, Phang, Pieler, Prashanth, Purohit, Reynolds, Tow, Wang,
  and Weinbach]{gpt_neox}
Sidney Black, Stella Biderman, Eric Hallahan, Quentin Anthony, Leo Gao,
  Laurence Golding, Horace He, Connor Leahy, Kyle McDonell, Jason Phang,
  Michael Pieler, Usvsn~Sai Prashanth, Shivanshu Purohit, Laria Reynolds,
  Jonathan Tow, Ben Wang, and Samuel Weinbach.
\newblock {GPT}-{N}eo{X}-20{B}: An open-source autoregressive language model.
\newblock In \emph{Proceedings of BigScience Episode {\#}5 -- Workshop on
  Challenges {\&} Perspectives in Creating Large Language Models}, pp.\
  95--136, virtual+Dublin, May 2022. Association for Computational Linguistics.
\newblock \doi{10.18653/v1/2022.bigscience-1.9}.
\newblock URL \url{https://aclanthology.org/2022.bigscience-1.9}.

\bibitem[Brown et~al.(2020)Brown, Mann, Ryder, Subbiah, Kaplan, Dhariwal,
  Neelakantan, Shyam, Sastry, Askell, Agarwal, Herbert-Voss, Krueger, Henighan,
  Child, Ramesh, Ziegler, Wu, Winter, Hesse, Chen, Sigler, Litwin, Gray, Chess,
  Clark, Berner, McCandlish, Radford, Sutskever, and Amodei]{gpt3}
Tom~B. Brown, Benjamin Mann, Nick Ryder, Melanie Subbiah, Jared Kaplan,
  Prafulla Dhariwal, Arvind Neelakantan, Pranav Shyam, Girish Sastry, Amanda
  Askell, Sandhini Agarwal, Ariel Herbert-Voss, Gretchen Krueger, T.~J.
  Henighan, Rewon Child, Aditya Ramesh, Daniel~M. Ziegler, Jeff Wu, Clemens
  Winter, Christopher Hesse, Mark Chen, Eric Sigler, Mateusz Litwin, Scott
  Gray, Benjamin Chess, Jack Clark, Christopher Berner, Sam McCandlish, Alec
  Radford, Ilya Sutskever, and Dario Amodei.
\newblock Language models are few-shot learners.
\newblock \emph{ArXiv}, abs/2005.14165, 2020.

\bibitem[Chandrasekaran et~al.(2017)Chandrasekaran, Yadav, Chattopadhyay,
  Prabhu, and Parikh]{theory_of_ais_mind}
Arjun Chandrasekaran, Deshraj Yadav, Prithvijit Chattopadhyay, Viraj Prabhu,
  and Devi Parikh.
\newblock It takes two to tango: Towards theory of ai's mind.
\newblock \emph{ArXiv}, abs/1704.00717, 2017.

\bibitem[Dennett(1991)]{dennett91}
Daniel~C. Dennett.
\newblock Two contrasts: Folk craft vs folk science and belief vs opinion.
\newblock In John~D. Greenwood (ed.), \emph{The Future of Folk Psychology},
  pp.\  135--148. Cambridge University Press, 1991.

\bibitem[Devlin et~al.(2019)Devlin, Chang, Lee, and Toutanova]{bert}
Jacob Devlin, Ming-Wei Chang, Kenton Lee, and Kristina Toutanova.
\newblock Bert: Pre-training of deep bidirectional transformers for language
  understanding.
\newblock In \emph{NAACL}, 2019.

\bibitem[Elazar et~al.(2021)Elazar, Kassner, Ravfogel, Ravichander, Hovy,
  Sch{\"u}tze, and Goldberg]{pararel}
Yanai Elazar, Nora Kassner, Shauli Ravfogel, Abhilasha Ravichander, Eduard~H.
  Hovy, Hinrich Sch{\"u}tze, and Yoav Goldberg.
\newblock Measuring and improving consistency in pretrained language models.
\newblock \emph{Transactions of the Association for Computational Linguistics},
  9:\penalty0 1012--1031, 2021.

\bibitem[Ettinger(2020)]{what_bert_is_not}
Allyson Ettinger.
\newblock What bert is not: Lessons from a new suite of psycholinguistic
  diagnostics for language models.
\newblock \emph{Transactions of the Association for Computational Linguistics},
  8:\penalty0 34--48, 2020.

\bibitem[Goyal et~al.(2019)Goyal, Feder, Shalit, and Kim]{cace}
Yash Goyal, Amir Feder, Uri Shalit, and Been Kim.
\newblock Explaining classifiers with causal concept effect (cace).
\newblock \emph{ArXiv}, abs/1907.07165, 2019.

\bibitem[Hase et~al.(2021)Hase, Diab, Celikyilmaz, Li, Kozareva, Stoyanov,
  Bansal, and Iyer]{lm_beliefs}
Peter Hase, Mona~T. Diab, Asli Celikyilmaz, Xian Li, Zornitsa Kozareva, Veselin
  Stoyanov, Mohit Bansal, and Srini Iyer.
\newblock Do language models have beliefs? methods for detecting, updating, and
  visualizing model beliefs.
\newblock \emph{ArXiv}, abs/2111.13654, 2021.

\bibitem[Hoffmann et~al.(2022)Hoffmann, Borgeaud, Mensch, Buchatskaya, Cai,
  Rutherford, de~Las~Casas, Hendricks, Welbl, Clark, Hennigan, Noland,
  Millican, van~den Driessche, Damoc, Guy, Osindero, Simonyan, Elsen, Rae,
  Vinyals, and Sifre]{gopher_scaling}
Jordan Hoffmann, Sebastian Borgeaud, Arthur Mensch, Elena Buchatskaya, Trevor
  Cai, Eliza Rutherford, Diego de~Las~Casas, Lisa~Anne Hendricks, Johannes
  Welbl, Aidan Clark, Tom Hennigan, Eric Noland, Katie Millican, George van~den
  Driessche, Bogdan Damoc, Aurelia Guy, Simon Osindero, Karen Simonyan, Erich
  Elsen, Jack~W. Rae, Oriol Vinyals, and L.~Sifre.
\newblock Training compute-optimal large language models.
\newblock \emph{ArXiv}, abs/2203.15556, 2022.

\bibitem[Jang et~al.(2021)Jang, Kwon, and Lukasiewicz]{calum}
Myeongjun Jang, Deuk~Sin Kwon, and Thomas Lukasiewicz.
\newblock Accurate, yet inconsistent? consistency analysis on language
  understanding models.
\newblock \emph{ArXiv}, abs/2108.06665, 2021.

\bibitem[Kassner \& Sch{\"u}tze(2020)Kassner and
  Sch{\"u}tze]{negated_misprimed}
Nora Kassner and Hinrich Sch{\"u}tze.
\newblock Negated and misprimed probes for pretrained language models: Birds
  can talk, but cannot fly.
\newblock In \emph{ACL}, 2020.

\bibitem[Kim et~al.(2018)Kim, Wattenberg, Gilmer, Cai, Wexler, Vi{\'e}gas, and
  Sayres]{tcav}
Been Kim, Martin Wattenberg, Justin Gilmer, Carrie~J. Cai, James Wexler,
  Fernanda~B. Vi{\'e}gas, and Rory Sayres.
\newblock Interpretability beyond feature attribution: Quantitative testing
  with concept activation vectors (tcav).
\newblock In \emph{ICML}, 2018.

\bibitem[Ma et~al.(2019)Ma, Francis, Lu, Nyberg, and Oltramari]{towards_gn}
Kaixin Ma, Jonathan Francis, Quanyang Lu, Eric Nyberg, and Alessandro
  Oltramari.
\newblock Towards generalizable neuro-symbolic systems for commonsense question
  answering.
\newblock \emph{ArXiv}, abs/1910.14087, 2019.

\bibitem[Piantadosi \& Hill(2022)Piantadosi and
  Hill]{meaning_without_reference}
Steven~T. Piantadosi and Felix Hill.
\newblock Meaning without reference in large language models.
\newblock \emph{ArXiv}, abs/2208.02957, 2022.

\bibitem[Premack \& Woodruff(1978)Premack and Woodruff]{premack_woodruff}
David Premack and Guy Woodruff.
\newblock Does the chimpanzee have a theory of mind?
\newblock \emph{Behavioral and Brain Sciences}, 1:\penalty0 515 -- 526, 1978.

\bibitem[Rabinowitz et~al.(2018)Rabinowitz, Perbet, Song, Zhang, Eslami, and
  Botvinick]{machine_tom}
Neil~C. Rabinowitz, Frank Perbet, H.~Francis Song, Chiyuan Zhang, S.~M.~Ali
  Eslami, and Matthew~M. Botvinick.
\newblock Machine theory of mind.
\newblock In \emph{ICML}, 2018.

\bibitem[Ray et~al.(2019)Ray, Sikka, Divakaran, Lee, and Burachas]{convqa}
Arijit Ray, Karan Sikka, Ajay Divakaran, Stefan Lee, and Giedrius Burachas.
\newblock Sunny and dark outside?! improving answer consistency in vqa through
  entailed question generation.
\newblock \emph{ArXiv}, abs/1909.04696, 2019.

\bibitem[Ribeiro et~al.(2019)Ribeiro, Guestrin, and Singh]{are_roses_red}
Marco~Tulio Ribeiro, Carlos Guestrin, and Sameer Singh.
\newblock Are red roses red? evaluating consistency of question-answering
  models.
\newblock In \emph{ACL}, 2019.

\bibitem[Ribeiro et~al.(2020)Ribeiro, Wu, Guestrin, and Singh]{checklist}
Marco~Tulio Ribeiro, Tongshuang~Sherry Wu, Carlos Guestrin, and Sameer Singh.
\newblock Beyond accuracy: Behavioral testing of nlp models with checklist.
\newblock In \emph{ACL}, 2020.

\bibitem[Sahu et~al.(2021)Sahu, Cogswell, Divakaran, and
  Rutherford-Quach]{sahu_bloom}
Pritish Sahu, Michael Cogswell, Ajay Divakaran, and Sara Rutherford-Quach.
\newblock Comprehension based question answering using bloom{'}s taxonomy.
\newblock In \emph{Proceedings of the 6th Workshop on Representation Learning
  for NLP (RepL4NLP-2021)}, pp.\  20--28, Online, August 2021. Association for
  Computational Linguistics.

\bibitem[Sanh et~al.(2022)Sanh, Webson, Raffel, Bach, Sutawika, Alyafeai,
  Chaffin, Stiegler, Scao, Raja, Dey, Bari, Xu, Thakker, Sharma, Szczechla,
  Kim, Chhablani, Nayak, Datta, Chang, Jiang, Wang, Manica, Shen, Yong, Pandey,
  Bawden, Wang, Neeraj, Rozen, Sharma, Santilli, F{\'e}vry, Fries, Teehan,
  Biderman, Gao, Bers, Wolf, and Rush]{t0}
Victor Sanh, Albert Webson, Colin Raffel, Stephen~H. Bach, Lintang~A. Sutawika,
  Zaid Alyafeai, Antoine Chaffin, Arnaud Stiegler, Teven~Le Scao, Arun Raja,
  Manan Dey, M~Saiful Bari, Canwen Xu, Urmish Thakker, Shanya Sharma, Eliza
  Szczechla, Taewoon Kim, Gunjan Chhablani, Nihal~V. Nayak, Debajyoti Datta,
  Jonathan Chang, Mike Tian-Jian Jiang, Han Wang, Matteo Manica, Sheng Shen,
  Zheng~Xin Yong, Harshit Pandey, Rachel Bawden, Thomas Wang, Trishala Neeraj,
  Jos Rozen, Abheesht Sharma, Andrea Santilli, Thibault F{\'e}vry, Jason~Alan
  Fries, Ryan Teehan, Stella~Rose Biderman, Leo Gao, Tali Bers, Thomas Wolf,
  and Alexander~M. Rush.
\newblock Multitask prompted training enables zero-shot task generalization.
\newblock \emph{ArXiv}, abs/2110.08207, 2022.

\bibitem[Shvo et~al.(2020)Shvo, Klassen, and McIlraith]{tom_explanation}
Maayan Shvo, Toryn~Q. Klassen, and Sheila~A. McIlraith.
\newblock Towards the role of theory of mind in explanation.
\newblock \emph{Explainable, Transparent Autonomous Agents and Multi-Agent
  Systems}, 12175:\penalty0 75 -- 93, 2020.

\bibitem[Shwartz et~al.(2020)Shwartz, West, Bras, Bhagavatula, and
  Choi]{vered_unsup}
Vered Shwartz, Peter West, Ronan~Le Bras, Chandra Bhagavatula, and Yejin Choi.
\newblock Unsupervised commonsense question answering with self-talk.
\newblock \emph{ArXiv}, abs/2004.05483, 2020.

\bibitem[Speer et~al.(2017)Speer, Chin, and Havasi]{conceptnet}
Robyn Speer, Joshua Chin, and Catherine Havasi.
\newblock Conceptnet 5.5: An open multilingual graph of general knowledge.
\newblock \emph{ArXiv}, abs/1612.03975, 2017.

\bibitem[Srivastava et~al.(2022)Srivastava, Rastogi, Rao, Shoeb, Abid, Fisch,
  Brown, Santoro, Gupta, Garriga-Alonso, Kluska, Lewkowycz, Agarwal, Power,
  Ray, Warstadt, Kocurek, Safaya, Tazarv, Xiang, Parrish, Nie, Hussain, Askell,
  Dsouza, Rahane, Iyer, Andreassen, Santilli, Stuhlmuller, Dai, La, Lampinen,
  Zou, Jiang, Chen, Vuong, Gupta, Gottardi, Norelli, Venkatesh, Gholamidavoodi,
  Tabassum, Menezes, Kirubarajan, Mullokandov, Sabharwal, Herrick, Efrat,
  Erdem, Karakacs, Roberts, Loe, Zoph, Bojanowski, Ozyurt, Hedayatnia,
  Neyshabur, Inden, Stein, Ekmekci, Lin, Howald, Diao, Dour, Stinson, Argueta,
  Ram'irez, Singh, Rathkopf, Meng, Baral, Wu, Callison-Burch, Waites, Voigt,
  Manning, Potts, Ramirez, Rivera, Siro, Raffel, Ashcraft, Garbacea, Sileo,
  Garrette, Hendrycks, Kilman, Roth, Freeman, Khashabi, Levy, Gonz'alez,
  Hernandez, Chen, Ippolito, Gilboa, Dohan, Drakard, Jurgens, Datta, Ganguli,
  Emelin, Kleyko, Yuret, Chen, Tam, Hupkes, Misra, Buzan, Mollo, Yang, Lee,
  Shutova, Cubuk, Segal, Hagerman, Barnes, Donoway, Pavlick, Rodol{\`a}, Lam,
  Chu, Tang, Erdem, Chang, Chi, Dyer, Jerzak, Kim, Manyasi, Zheltonozhskii,
  Xia, Siar, Mart'inez-Plumed, Happ'e, Chollet, Rong, Mishra, Winata, de~Melo,
  Kruszewski, Parascandolo, Mariani, Wang, Jaimovitch-L'opez, Betz, Gur-Ari,
  Galijasevic, Kim, Rashkin, Hajishirzi, Mehta, Bogar, Shevlin, Sch{\"u}tze,
  Yakura, Zhang, Wong, Ng, Noble, Jumelet, Geissinger, Kernion, Hilton, Lee,
  Fisac, Simon, Koppel, Zheng, Zou, Koco'n, Thompson, Kaplan, Radom,
  Sohl-Dickstein, Phang, Wei, Yosinski, Novikova, Bosscher, Marsh, Kim, Taal,
  Engel, Alabi, Xu, Song, Tang, Waweru, Burden, Miller, Balis, Berant,
  Frohberg, Rozen, Hern{\'a}ndez-Orallo, Boudeman, Jones, Tenenbaum, Rule,
  Chua, Kanclerz, Livescu, Krauth, Gopalakrishnan, Ignatyeva, Markert, Dhole,
  Gimpel, Omondi, Mathewson, Chiafullo, Shkaruta, Shridhar, McDonell,
  Richardson, Reynolds, Gao, Zhang, Dugan, Qin, Contreras-Ochando, Morency,
  Moschella, Lam, Noble, Schmidt, He, Col'on, Metz, cSenel, Bosma, Sap, ter
  Hoeve, Andrea, Farooqi, Faruqui, Mazeika, Baturan, Marelli, Maru, Quintana,
  Tolkiehn, Giulianelli, Lewis, Potthast, Leavitt, Hagen, Schubert,
  Baitemirova, Arnaud, McElrath, Yee, Cohen, Gu, Ivanitskiy, Starritt, Strube,
  Swkedrowski, Bevilacqua, Yasunaga, Kale, Cain, Xu, Suzgun, Tiwari, Bansal,
  Aminnaseri, Geva, Gheini, MukundVarma, Peng, Chi, Lee, Krakover, Cameron,
  Roberts, Doiron, Nangia, Deckers, Muennighoff, Keskar, Iyer, Constant,
  Fiedel, Wen, Zhang, Agha, Elbaghdadi, Levy, Evans, Casares, Doshi, Fung,
  Liang, Vicol, Alipoormolabashi, Liao, Liang, Chang, Eckersley, Htut, Hwang,
  Milkowski, Patil, Pezeshkpour, Oli, Mei, LYU, Chen, Banjade, Rudolph,
  Gabriel, Habacker, Delgado, Milli{\`e}re, Garg, Barnes, Saurous, Arakawa,
  Raymaekers, Frank, Sikand, Novak, Sitelew, Bras, Liu, Jacobs, Zhang,
  Salakhutdinov, Chi, Lee, Stovall, Teehan, Yang, Singh, Mohammad, Anand,
  Dillavou, Shleifer, Wiseman, Gruetter, Bowman, Schoenholz, Han, Kwatra, Rous,
  Ghazarian, Ghosh, Casey, Bischoff, Gehrmann, Schuster, Sadeghi, Hamdan, Zhou,
  Srivastava, Shi, Singh, Asaadi, Gu, Pachchigar, Toshniwal, Upadhyay, Debnath,
  Shakeri, Thormeyer, Melzi, Reddy, Makini, hwan Lee, Torene, Hatwar, Dehaene,
  Divic, Ermon, Biderman, Lin, Prasad, Piantadosi, Shieber, Misherghi,
  Kiritchenko, Mishra, Linzen, Schuster, Li, Yu, Ali, Hashimoto, Wu, Desbordes,
  Rothschild, Phan, Wang, Nkinyili, Schick, Kornev, Telleen-Lawton, Tunduny,
  Gerstenberg, Chang, Neeraj, Khot, Shultz, Shaham, Misra, Demberg, Nyamai,
  Raunak, Ramasesh, Prabhu, Padmakumar, Srikumar, Fedus, Saunders, Zhang,
  Vossen, Ren, Tong, Wu, Shen, Yaghoobzadeh, Lakretz, Song, Bahri, Choi, Yang,
  Hao, Chen, Belinkov, Hou, Hou, Bai, Seid, Xinran, Zhao, Wang, Wang, Wang, Wu,
  Singh, and Shaham]{big_bench}
Aarohi Srivastava, Abhinav Rastogi, Abhishek~B Rao, Abu Awal~Md Shoeb, Abubakar
  Abid, Adam Fisch, Adam~R. Brown, Adam Santoro, Aditya Gupta, Adri{\`a}
  Garriga-Alonso, Agnieszka Kluska, Aitor Lewkowycz, Akshat Agarwal, Alethea
  Power, Alex Ray, Alex Warstadt, Alexander~W. Kocurek, Ali Safaya, Ali Tazarv,
  Alice Xiang, Alicia Parrish, Allen Nie, Aman Hussain, Amanda Askell, Amanda
  Dsouza, Ameet~Annasaheb Rahane, Anantharaman~S. Iyer, Anders~Johan
  Andreassen, Andrea Santilli, Andreas Stuhlmuller, Andrew~M. Dai, Andrew~D.
  La, Andrew~Kyle Lampinen, Andy Zou, Angela Jiang, Angelica Chen, Anh Vuong,
  Animesh Gupta, Anna Gottardi, Antonio Norelli, Anu Venkatesh, Arash
  Gholamidavoodi, Arfa Tabassum, Arul Menezes, Arun Kirubarajan, Asher
  Mullokandov, Ashish Sabharwal, Austin Herrick, Avia Efrat, Aykut Erdem, Ayla
  Karakacs, Bridget~R. Roberts, Bao~Sheng Loe, Barret Zoph, Bartlomiej
  Bojanowski, Batuhan Ozyurt, Behnam Hedayatnia, Behnam Neyshabur, Benjamin
  Inden, Benno Stein, Berk Ekmekci, Bill~Yuchen Lin, Blake~Stephen Howald,
  Cameron Diao, Cameron Dour, Catherine Stinson, Cedrick Argueta, C'esar~Ferri
  Ram'irez, Chandan Singh, Charles Rathkopf, Chenlin Meng, Chitta Baral, Chiyu
  Wu, Chris Callison-Burch, Chris Waites, Christian Voigt, Christopher~D.
  Manning, Christopher Potts, Cindy~Tatiana Ramirez, Clara Rivera, Clemencia
  Siro, Colin Raffel, Courtney Ashcraft, Cristina Garbacea, Damien Sileo,
  Daniel~H Garrette, Dan Hendrycks, Dan Kilman, Dan Roth, Daniel Freeman,
  Daniel Khashabi, Daniel Levy, Daniel Gonz'alez, Danny Hernandez, Danqi Chen,
  Daphne Ippolito, Dar Gilboa, David Dohan, D.~Drakard, David Jurgens,
  Debajyoti Datta, Deep Ganguli, Denis Emelin, Denis Kleyko, Deniz Yuret, Derek
  Chen, Derek Tam, Dieuwke Hupkes, Diganta Misra, Dilyar Buzan, Dimitri~Coelho
  Mollo, Diyi Yang, Dong-Ho Lee, Ekaterina Shutova, Ekin~Dogus Cubuk, Elad
  Segal, Eleanor Hagerman, Elizabeth Barnes, Elizabeth~P. Donoway, Ellie
  Pavlick, Emanuele Rodol{\`a}, Emma~FC Lam, Eric Chu, Eric Tang, Erkut Erdem,
  Ernie Chang, Ethan~A. Chi, Ethan Dyer, Ethan Jerzak, Ethan Kim, Eunice~Engefu
  Manyasi, Evgenii Zheltonozhskii, Fan Xia, Fatemeh Siar, Fernando
  Mart'inez-Plumed, Francesca Happ'e, François Chollet, Frieda Rong, Gaurav
  Mishra, Genta~Indra Winata, Gerard de~Melo, Germ{\'a}n Kruszewski,
  Giambattista Parascandolo, Giorgio Mariani, Gloria Wang, Gonzalo
  Jaimovitch-L'opez, Gregor Betz, Guy Gur-Ari, Hana Galijasevic, Han~Sol Kim,
  Hannah Rashkin, Hanna Hajishirzi, Harsh Mehta, Hayden Bogar, Henry Shevlin,
  Hinrich Sch{\"u}tze, Hiromu Yakura, Hongming Zhang, Hubert Wong, Ian Aik-Soon
  Ng, Isaac Noble, Jaap Jumelet, Jack Geissinger, John Kernion, Jacob Hilton,
  Jaehoon Lee, Jaime~Fern{\'a}ndez Fisac, J.~Brooker Simon, James Koppel, James
  Zheng, James Zou, Jan Koco'n, Jana Thompson, Jared Kaplan, Jarema Radom,
  Jascha~Narain Sohl-Dickstein, Jason Phang, Jason Wei, Jason Yosinski,
  Jekaterina Novikova, Jelle Bosscher, Jenni Marsh, Jeremy Kim, Jeroen Taal,
  Jesse Engel, Jesujoba~Oluwadara Alabi, Jiacheng Xu, Jiaming Song, Jillian
  Tang, Jane~W Waweru, John Burden, John Miller, John~U. Balis, Jonathan
  Berant, Jorg Frohberg, Jos Rozen, Jos{\'e} Hern{\'a}ndez-Orallo, Joseph
  Boudeman, Joseph Jones, Joshua~B. Tenenbaum, Joshua~S. Rule, Joyce Chua,
  Kamil Kanclerz, Karen Livescu, Karl Krauth, Karthik Gopalakrishnan, Katerina
  Ignatyeva, Katja Markert, Kaustubh~D. Dhole, Kevin Gimpel, Kevin~Ochieng’
  Omondi, Kory~Wallace Mathewson, Kristen Chiafullo, Ksenia Shkaruta, Kumar
  Shridhar, Kyle McDonell, Kyle Richardson, Laria Reynolds, Leo Gao, Li~Zhang,
  Liam Dugan, Lianhui Qin, Lidia Contreras-Ochando, Louis-Philippe Morency,
  Luca Moschella, Luca Lam, Lucy Noble, Ludwig Schmidt, Luheng He,
  Luis~Oliveros Col'on, Luke Metz, Lutfi~Kerem cSenel, Maarten Bosma, Maarten
  Sap, Maartje ter Hoeve, Madotto Andrea, Maheen~Saleem Farooqi, Manaal
  Faruqui, Mantas Mazeika, Marco Baturan, Marco Marelli, Marco Maru,
  M~Quintana, Marie Tolkiehn, Mario Giulianelli, Martha Lewis, Martin Potthast,
  Matthew Leavitt, Matthias Hagen, M'aty'as Schubert, Medina Baitemirova,
  Melissa Arnaud, Melvin~Andrew McElrath, Michael~A. Yee, Michael Cohen, Mi~Gu,
  Michael~I. Ivanitskiy, Michael Starritt, Michael Strube, Michal Swkedrowski,
  Michele Bevilacqua, Michihiro Yasunaga, Mihir Kale, Mike Cain, Mimee Xu,
  Mirac Suzgun, Monica Tiwari, Mohit Bansal, Moin Aminnaseri, Mor Geva, Mozhdeh
  Gheini, T~MukundVarma, Nanyun Peng, Nathan Chi, Nayeon Lee, Neta Gur-Ari
  Krakover, Nicholas Cameron, Nicholas~S. Roberts, Nicholas Doiron, Nikita
  Nangia, Niklas Deckers, Niklas Muennighoff, Nitish~Shirish Keskar, Niveditha
  Iyer, Noah Constant, Noah Fiedel, Nuan Wen, Oliver Zhang, Omar Agha, Omar
  Elbaghdadi, Omer Levy, Owain Evans, Pablo Antonio~Moreno Casares, Parth
  Doshi, Pascale Fung, Paul~Pu Liang, Paul Vicol, Pegah Alipoormolabashi,
  Peiyuan Liao, Percy Liang, Peter~W. Chang, Peter Eckersley, Phu~Mon Htut,
  Pi-Bei Hwang, P.~Milkowski, Piyush~S. Patil, Pouya Pezeshkpour, Priti Oli,
  Qiaozhu Mei, QING LYU, Qinlang Chen, Rabin Banjade, Rachel~Etta Rudolph,
  Raefer Gabriel, Rahel Habacker, Ram'on~Risco Delgado, Rapha{\"e}l
  Milli{\`e}re, Rhythm Garg, Richard Barnes, Rif~A. Saurous, Riku Arakawa,
  Robbe Raymaekers, Robert Frank, Rohan Sikand, Roman Novak, Roman Sitelew,
  Ronan~Le Bras, Rosanne Liu, Rowan Jacobs, Rui Zhang, Ruslan Salakhutdinov,
  Ryan Chi, Ryan Lee, Ryan Stovall, Ryan Teehan, Rylan Yang, Sahib~J. Singh,
  Saif~M. Mohammad, Sajant Anand, Sam Dillavou, Sam Shleifer, Sam Wiseman,
  Samuel Gruetter, Sam Bowman, Samuel~S. Schoenholz, Sanghyun Han, Sanjeev
  Kwatra, Sarah~A. Rous, Sarik Ghazarian, Sayan Ghosh, Sean Casey, Sebastian
  Bischoff, Sebastian Gehrmann, Sebastian Schuster, Sepideh Sadeghi, Shadi~S.
  Hamdan, Sharon Zhou, Shashank Srivastava, Sherry Shi, Shikhar Singh, Shima
  Asaadi, Shixiang~Shane Gu, Shubh Pachchigar, Shubham Toshniwal, Shyam
  Upadhyay, Shyamolima Debnath, Siamak Shakeri, Simon Thormeyer, Simone Melzi,
  Siva Reddy, Sneha~Priscilla Makini, Soo hwan Lee, Spencer~Bradley Torene,
  Sriharsha Hatwar, Stanislas Dehaene, Stefan Divic, Stefano Ermon, Stella~Rose
  Biderman, Stephanie~C. Lin, Stephen Prasad, Steven~T. Piantadosi, Stuart~M.
  Shieber, Summer Misherghi, Svetlana Kiritchenko, Swaroop Mishra, Tal Linzen,
  Tal Schuster, Tao Li, Tao Yu, Tariq~A. Ali, Tatsuo Hashimoto, Te-Lin Wu, Theo
  Desbordes, Theodore Rothschild, Thomas Phan, Tianle Wang, Tiberius Nkinyili,
  Timo Schick, T.~N. Kornev, Timothy Telleen-Lawton, Titus Tunduny, Tobias
  Gerstenberg, Trenton Chang, Trishala Neeraj, Tushar Khot, Tyler~O. Shultz,
  Uri Shaham, Vedant Misra, Vera Demberg, Victoria Nyamai, Vikas Raunak,
  Vinay~Venkatesh Ramasesh, Vinay~Uday Prabhu, Vishakh Padmakumar, Vivek
  Srikumar, William Fedus, William Saunders, William Zhang, W~Vossen, Xiang
  Ren, Xiaoyu~F Tong, Xinyi Wu, Xudong Shen, Yadollah Yaghoobzadeh, Yair
  Lakretz, Yang Song, Yasaman Bahri, Ye~Ji Choi, Yichi Yang, Yiding Hao, Yifu
  Chen, Yonatan Belinkov, Yu~Hou, Yu~Hou, Yushi Bai, Zachary Seid, Zhao Xinran,
  Zhuoye Zhao, Zi~Fu Wang, Zijie~J. Wang, Zirui Wang, Ziyi Wu, Sahib Singh, and
  Uri Shaham.
\newblock Beyond the imitation game: Quantifying and extrapolating the
  capabilities of language models.
\newblock \emph{ArXiv}, abs/2206.04615, 2022.

\bibitem[Talmor et~al.(2019)Talmor, Herzig, Lourie, and
  Berant]{talmor2019commonsenseqa}
Alon Talmor, Jonathan Herzig, Nicholas Lourie, and Jonathan Berant.
\newblock Commonsenseqa: A question answering challenge targeting commonsense
  knowledge.
\newblock In \emph{Proceedings of the 2019 Conference of the North American
  Chapter of the Association for Computational Linguistics: Human Language
  Technologies, Volume 1 (Long and Short Papers)}, pp.\  4149--4158, 2019.

\bibitem[Talmor et~al.(2021)Talmor, Yoran, Bras, Bhagavatula, Goldberg, Choi,
  and Berant]{csqa2}
Alon Talmor, Ori Yoran, Ronan~Le Bras, Chandrasekhar Bhagavatula, Yoav
  Goldberg, Yejin Choi, and Jonathan Berant.
\newblock Commonsenseqa 2.0: Exposing the limits of ai through gamification.
\newblock In \emph{NeurIPS Datasets and Benchmarks}, 2021.

\bibitem[Wang et~al.(2022)Wang, Wei, Schuurmans, Le, Chi, and
  Zhou]{self_chain_of_thought}
Xuezhi Wang, Jason Wei, Dale Schuurmans, Quoc Le, Ed~Chi, and Denny Zhou.
\newblock Self-consistency improves chain of thought reasoning in language
  models.
\newblock \emph{ArXiv}, abs/2203.11171, 2022.

\bibitem[Wei et~al.(2022{\natexlab{a}})Wei, Tay, Bommasani, Raffel, Zoph,
  Borgeaud, Yogatama, Bosma, Zhou, Metzler, Chi, Hashimoto, Vinyals, Liang,
  Dean, and Fedus]{emergent_llm}
Jason Wei, Yi~Tay, Rishi Bommasani, Colin Raffel, Barret Zoph, Sebastian
  Borgeaud, Dani Yogatama, Maarten Bosma, Denny Zhou, Donald Metzler, Ed~Chi,
  Tatsunori Hashimoto, Oriol Vinyals, Percy Liang, Jeff Dean, and William
  Fedus.
\newblock Emergent abilities of large language models.
\newblock \emph{ArXiv}, abs/2206.07682, 2022{\natexlab{a}}.

\bibitem[Wei et~al.(2022{\natexlab{b}})Wei, Wang, Schuurmans, Bosma, Chi, Le,
  and Zhou]{reason_chain_of_thought}
Jason Wei, Xuezhi Wang, Dale Schuurmans, Maarten Bosma, Ed~Chi, Quoc Le, and
  Denny Zhou.
\newblock Chain of thought prompting elicits reasoning in large language
  models.
\newblock \emph{ArXiv}, abs/2201.11903, 2022{\natexlab{b}}.

\bibitem[Wu et~al.(2021)Wu, Ribeiro, Heer, and Weld]{polyjuice}
Tongshuang~Sherry Wu, Marco~Tulio Ribeiro, Jeffrey Heer, and Daniel~S. Weld.
\newblock Polyjuice: Generating counterfactuals for explaining, evaluating, and
  improving models.
\newblock In \emph{ACL}, 2021.

\bibitem[Yeh et~al.(2021)Yeh, Kim, and Ravikumar]{concept_explanations}
Chih-Kuan Yeh, Been Kim, and Pradeep Ravikumar.
\newblock Human-centered concept explanations for neural networks.
\newblock In \emph{Neuro-Symbolic Artificial Intelligence}, 2021.

\bibitem[Zhang et~al.(2022{\natexlab{a}})Zhang, Li, Meng, Chang, and den
  Broeck]{paradox_reason_from_data}
Honghua Zhang, Liunian~Harold Li, Tao Meng, Kai-Wei Chang, and Guy~Van den
  Broeck.
\newblock On the paradox of learning to reason from data.
\newblock \emph{ArXiv}, abs/2205.11502, 2022{\natexlab{a}}.

\bibitem[Zhang et~al.(2022{\natexlab{b}})Zhang, Roller, Goyal, Artetxe, Chen,
  Chen, Dewan, Diab, Li, Lin, Mihaylov, Ott, Shleifer, Shuster, Simig, Koura,
  Sridhar, Wang, and Zettlemoyer]{opt}
Susan Zhang, Stephen Roller, Naman Goyal, Mikel Artetxe, Moya Chen, Shuohui
  Chen, Christopher Dewan, Mona Diab, Xian Li, Xi~Victoria Lin, Todor Mihaylov,
  Myle Ott, Sam Shleifer, Kurt Shuster, Daniel Simig, Punit~Singh Koura, Anjali
  Sridhar, Tianlu Wang, and Luke Zettlemoyer.
\newblock Opt: Open pre-trained transformer language models.
\newblock \emph{ArXiv}, abs/2205.01068, 2022{\natexlab{b}}.

\end{thebibliography}
\bibliographystyle{iclr2023_conference}

\newpage
\appendix
\section{Appendix}
We prepared different prompt templates for the picked 14 relations used in our experiments shown in \autoref{tab:prompt_style}. 
\begin{table*}[!ht]
\caption{All the relations with its corresponding prompt style with a sample example.}
\centering
\label{tab:prompt_style}
\begin{tabular}{lll} 
\hline\noalign{\smallskip}
\textbf{Relation}   & \textbf{Prompt Style} & \textbf{Sample Instance} \\ 
\hline \noalign{\smallskip}
Is A & Is $c^1$ a $c^2$? & Is security a department? \\
Has A & Does $c^1$ has a $c^2$? & does house has a basement? \\
Antonym & Is $c^1$ an antonym of $c^2$? & Is clever an antonym of dull? \\
Cause & Does $c^1$ cause $c^2$? & does fencing cause small cuts? \\
Desires & Does a $c^1$ desires $c^2$? & does a dog desires affection? \\
Form Of & Is $c^1$ a form of $c^2$? & Is partying a form of of party? \\
Made Of & Is the $c^1$ made of $c^2$? & Is the car made of metal? \\
Part Of & Is $c^1$ a part of $c^2$? & Is book a part of library? \\
Related To & Is $c^1$ related to $c^2$? & Is doctor related to illness? \\
Similar To & Is $c^1$ similar to $c^2$? & Is ridiculous similar to silly? \\
Synonym & Is $c^1$ a synonym of $c^2$? & Is reply a synonym of answer? \\
Used For & Are $c^1$ used for $c^2$? & Are clothes used for wearing? \\
At Location & Is $c^1$ at location $c^2$? & Is door at location library? \\
Capable Of & Is a $c^1$ capable of $c^2$? & Is a child capable of form opinions? \\
\end{tabular}
\end{table*}

In \autoref{tab:neg_tuples}, we showcase few instances of the negative background facts created from the process described in \autoref{sec:bk_extract}.

\begin{table*}[!ht]
\caption{Examples of negative background knowledge task for each relation.}
\centering
\label{tab:neg_tuples}
\resizebox{0.99\columnwidth}{!}{%
\begin{tabular}{ll} 
\hline\noalign{\smallskip}
\textbf{Relation}   & \textbf{Negative Background Facts}\\ 
\hline \noalign{\smallskip}
Is A &  [Is drill a clamp?, Is ocean a shame?, Is space a micrometer?] \\
Has A &  [does mammals has a watch?, does pen has a unicycle?, does oceans has a uncle?] \\
Antonym &  [Is wash an antonym of detached?, Is bad an antonym of nightdress?, Is shade an antonym of improvement?]\\
Cause &  [does going into coma cause company?, does standing in queue cause shrinking?, does doing housework cause mermaid?] \\
Desires &  [does a person desires schizophrenia?, does a children desires unemployed?, does a tree desires criticism?] \\
Form Of &  [Is recycled a form of burned?, Is bath room a form of interrupted?, Is storing a form of bleeding?] \\
Made Of &  [Is the ocean made of uncomfortableness?, Is the car made of rain?, Is the plants made of bicycle?] \\
Part Of &  [Is gulf a part of round?, Is shower a part of anger?, Is bell a part of congress?] \\
Related To &  [Is class related to cornstarch?, Is room related to jordan?, Is cable related to tilemaking?] \\
Similar To &  [Is lie similar to botany?, Is ridiculous similar to aspirin?, Is distant similar to physiology?] \\
Synonym &  [Is heart a synonym of volition?, Is station a synonym of subordination?, Is part a synonym of undermine?] \\
Used For &  [Are hair used for council?, Are theatre used for desk?, Are disk used for pain?] \\
At Location &  [Is monkey at location fuzzball?, Is piano at location macaroni?, Is table at location gauging?] \\
Capable Of &  [Is a computer capable of pillow?, Is a band capable of overmodulation?, Is a tiger capable of fireman?] \\
\end{tabular}
}
\end{table*}



\end{document}